\documentclass[3p]{elsarticle}

\usepackage{xr}
\usepackage{hyperref,array,amsmath,booktabs,amssymb,float, subfig, tabularx, multirow, appendix}
\usepackage{etoolbox}
\usepackage{tikz}

\newrobustcmd*{\triangled}[1]{\tikz{\filldraw[draw=#1,fill=#1] (0,0) --
(0.2cm,0) -- (0.1cm,-0.2cm);}$\,$}
\newrobustcmd*{\triangleu}[1]{\tikz{\filldraw[draw=#1,fill=#1] (0,0) --
(0.2cm,0) -- (0.1cm,0.2cm);}$\,$}

\usepackage{color}
\let\today\relax
\makeatletter
\def\ps@pprintTitle{%
    \let\@oddhead\@empty
    \let\@evenhead\@empty
    \def\@oddfoot{\footnotesize\itshape
         {Submitted preprint} \hfill\today}%
    \let\@evenfoot\@oddfoot
    }
\makeatother
\begin{document}

\begin{frontmatter}

\title{Robust Optimization and Validation of Echo State Networks \\for learning chaotic dynamics}

\author[1]{Alberto Racca}
\author[1,2,3]{Luca Magri\corref{cor1}}
\ead{lm547@cam.ac.uk}

\cortext[cor1]{Corresponding author}
\address[1]{Department of Engineering, University of Cambridge, Trumpington Street, Cambridge CB2 1PZ, UK}
\address[2]{The Alan Turing Institute, 96 Euston Road, London, England, NW1 2DB, UK}
\address[3]{Institute for Advanced Study, Technical University of Munich, Lichtenbergstrasse 2a, 85748 Garching,
Germany (visiting fellowship)}

\begin{abstract}
The time-accurate prediction of a chaotic system is challenging because its evolution becomes unpredictable after the predictability time.
This is because infinitesimal errors in a chaotic system increase exponentially, i.e., two nearby time series diverge from each other.
An approach to the  time-accurate prediction of chaotic solutions is by learning temporal patterns from data.
Echo State Networks (ESNs), which are a class of Reservoir Computing, can accurately predict the chaotic dynamics well beyond the predictability time.
Existing studies, however, also showed that small changes in the hyperparameters may markedly affect the network's performance. 
The overarching aim of this paper is to assess and improve the robustness of Echo State Networks for the time-accurate prediction of chaotic solutions.
The goal is three-fold.
First, we investigate the robustness of routinely used validation strategies.
Second, we propose the {\it Recycle Validation}, and the {\it chaotic versions} of existing validation strategies, to specifically tackle the forecasting of chaotic systems.
Third, we compare Bayesian optimization with the traditional Grid Search for optimal hyperparameter selection. 
Numerical tests are performed on two prototypical nonlinear systems that have both chaotic and quasiperiodic solutions.
Both model-free and model-informed Echo State Networks are analysed.
By comparing the network's robustness in learning  chaotic (unpredictable) versus quasiperiodic (predictable) solutions, we highlight fundamental challenges  in learning chaotic solutions. 

The proposed validation strategies, which are based on the dynamical systems properties of chaotic time series,  are shown to outperform the state-of-the-art validation strategies.
Because the strategies are principled---they are based on chaos theory such as the Lyapunov time---they can be  applied to other Recurrent Neural Networks architectures with little modification.
This work opens up new possibilities for the robust design and application of Echo State Networks, and Recurrent Neural Networks, to the time-accurate prediction of chaotic systems.
\end{abstract}


\end{frontmatter}

\section{Introduction}

Chaotic systems naturally appear in many branches of science and engineering, 
from turbulent flows~\cite[e.g.,][]{Deissler1986, Boffetta2002,Bec2006}, through vibrations~\cite{moon1983chaotic}, electronics and telecommunications~\cite{kennedy2000chaotic}, quantum mechanics~\cite{stockmann2000quantum}, reacting flows~\cite{Nastac2017,Hassanaly2019}, 
to epidemic modelling~\cite{bolker1993chaos}, to name only a few. 
The time-accurate computation of chaotic systems is hindered by the ``butterfly effect''~\cite{lorenz1963deterministic}:
an error in the system's knowledge---e.g, initial conditions and parameters---grows exponentially until nonlinear saturation. 
Practically,  it is not possible to time-accurately predict chaotic solutions after a time scale, known as the predictability time. 
The predictability time scales with the inverse of the dominant Lyapunov exponent, which is typically a small characteristic scale of the system under investigation~\cite{Boffetta2002}. 

%
An approach to the prediction of chaotic dynamics is data-driven. 
Given a time series (data), we wish to learn the underlying chaotic dynamics to predict the future evolution. 
The data-driven approach, also known as model-free, traces back to the delay coordinate embedding by Takens~\cite{takens1981detecting}, 
which is widely used in time series analysis, in particular, in low-dimensional systems~\cite{guckenheimer2013nonlinear}. 
An alternative data-driven approach to inferring (or, equivalently, learning) chaotic dynamics from data is machine learning. 
Machine learning is establishing itself as a paradigm that is complementary to first-principles modelling of nonlinear systems in computational  science and engineering~\cite{baker2019workshop}. 
%
%
In the realm of neural networks, which is the focus of this paper, the feed-forward neural network is the archetypical architecture, which may excel at classification and regression problems \cite{goodfellow2016deep}. 
The feed-forward neural network, however, is not the optimal architecture for chaotic time series forecasting because it not designed to learn temporal correlations. 
Specifically, in time series forecasting, inputs and outputs are ordered sequentially, in other words, they are temporally correlated. 
To overcome the limitations of feed-forward neural networks, Recurrent Neural Networks (RNNs) \cite{rumelhart1986learning} have been designed to learn temporal correlations. 
Examples of successful applications span from speech recognition \cite{sak2014long}, through language translation \cite{sutskever2014sequence}, 
fluids~\cite{brunton2020machine,nakai2018machine,wan2018machine,doan2019aphysics,vlachas2018data}, to thermo-acoustic oscillations \cite{huhn2020learning}, among many others. 
RNNs take into account the sequential nature of the inputs by updating a hidden time-varying state through an internal loop. 
As a result of the long-lasting time dependencies of the hidden state, however, training RNNs with Back Propagation Through Time \cite{werbos1990backpropagation} is notoriously difficult. 
This is because the repeated backwards multiplication of intermediate gradients cause the final gradient to either vanish or become unbounded depending on the spectral radius of the gradient matrix \cite{werbos1988generalization,bengio1994learning}. 
This makes the training ill-posed, which may negatively affect the computational of the optimal set of weights. 
To overcome this problem, two main types of RNN architectures have been proposed: Gated Structures and Reservoir Computing. 
Gated Structures prevent gradients from vanishing or becoming unbounded by regularizing the passage of information inside the network, as 
accomplished in architectures such as  Long Short-Term Memory (LSTM) networks \cite{hochreiter1997long} and  Gated Recurrent Units (GRU) networks \cite{cho2014learning}. 
Alternatively, in Reservoir Computing (RC)~\cite{jaeger2004harnessing,maass2002real}, a high-dimensional dynamical system, the reservoir, acts both as a nonlinear expansion of the inputs and as the memory of the system \cite{lukovsevivcius2012practical}. 
At each time step, the output is computed as a linear combination of the reservoir state's components, the weights of which are the only trainable parameters of the machine. 
Training is, therefore, reduced to a linear regression problem, which bypasses the issue of repeated gradients multiplication in RNNs.

In chaotic attractors, Reservoir Computing has been employed to achieve at least four different goals: 
to 
(i) learn ergodic properties, such as Lyapunov exponents~\cite{lu2018attractor,pathak2017using} and statistics \cite{lu2018attractor, huhn2020learning}; 
(ii) filter out noise to recover the deterministic dynamics~\cite{DOAN2020}, 
(iii) reconstruct unmeasured (hidden) variables \cite{lu2017reservoir,doan2020learning,racca2020automatic}
and 
(iv) time-accurately predict the dynamics~\cite{pathak2018model,doan2019physics,wikner2020combining}. 
In this work, we focus on the time-accurate short term prediction of chaotic attractors.
%
A successful Reservoir Computing architecture is the Echo State Network (ESN) \cite{jaeger2004harnessing}, which is a universal approximator~\cite{GRIGORYEVA2018495,GONON2021} suitable for the prediction of chaotic time series~\cite{pathak2018model}. 
There are two broad categories of Echo State Networks: model-free~\cite{pathak2018model,lukovsevivcius2012practical} and model-informed~\cite{pathak2018hybrid,doan2019physics}. 
On the one hand, in model-free ESNs, which are the original networks, the training is performed on data only~\cite{lukovsevivcius2012practical}. 
On the other hand, in model-informed ESNs,  the governing equations, or a reduced-order form of them, are embedded in the architecture, for example, in the reservoir in hybrid ESNs~\cite{pathak2018hybrid}, or in the loss function in physics-informed ESNs~\cite{doan2019physics} .
In chaotic time series forecasting, model-informed ESNs typically outperform model-free ESNs~\cite{pathak2018hybrid,doan2019physics, wikner2020combining}.
Both model-free and model-informed Echo State Networks perform as well as LSTMs and GRUs, requiring less computational resources for training \cite{vlachas2020backpropagation,chattopadhyay2019data}. 
 The robustness of ESNs for chaotic time series, however, has not been fully investigated yet, which motivates the overarching objective of this study. 
Two key aspects may affect ESN robustness. 
The first aspect is random the initialization, which is required to create the reservoir \cite{lukovsevivcius2012practical}.
Networks with different initializations may perform substantially differently, even after hyperparameter tuning \cite{haluszczynski2019good}. 
For an ESN to be robust, network testing through an ensemble of network realizations is required. 
The second aspect is  high hyperparameter sensitivity \cite{lukovsevivcius2012practical, jiang2019model}. 
The most common validation strategy to compute the hyperparameters for learning chaotic dynamics is the Single Shot Validation, which minimizes the error in an interval subsequent to the training interval. 
Other validation strategies have been investigated, such as the Walk Forward Validation and the K-Fold cross Validation~\cite{lukovsevivcius2019efficient}, but this study was restricted to non-chaotic systems. 
The computation of the optimal set of hyperparameters is typically performed by Grid Search \cite{jaeger2004harnessing,doan2019aphysics,pathak2018hybrid, pathak2018model}, 
although other optimization strategies such as Evolutionary Algorithms \cite{ishu2004identification, ferreira2013approach}, Stochastic Gradient Descent \cite{thiede2019gradient}, Particle Swarm Optimization \cite{wang2015optimizing} and Bayesian Optimization \cite{yperman2016bayesian} have been proposed. 
In particular, Bayesian Optimization (BO) has proved to improve the performance of reservoir-computing architectures in the prediction of chaotic time series,  outperforming the commonly used Grid Search strategy \cite{griffith2019forecasting}.
Bayesian Optimization is a gradient-free search strategy, thereby, it is less sensitive to local minima with respect to gradient descent methods \cite{yperman2016bayesian,thiede2019gradient}. 
Moreover, Bayesian Optimization is based on Gaussian Process (GP) regression \cite{rasmussen2003gaussian}, therefore, it naturally quantifies the uncertainty on the computation. \\

The objective of this paper is three-fold with a focus on learning chaotic dynamics from data. 
First, we investigate the robustness of the Single Shot Validation, Walk Forward Validation and the K-Fold cross Validation. 
Second, we propose the Recycle Validation and the chaotic version of existing validation strategies to specifically tackle the forecasting of chaotic systems. 
Third, we analyse Bayesian optimization for optimal hyperparameter selection.
The Lorenz system~\cite{lorenz1963deterministic} and the Kutznetsov oscillator~\cite{kuznetsov2010simple} are considered as prototypical low-order nonlinear deterministic systems. 
We highlight fundamental challenges in the robustness of ESNs for chaotic solutions with a comparative investigation on quasiperiodic oscillations. 
Both model-free and model-informed architectures are analysed. \\

The paper is organized as follows. 
Section~\ref{sec:ESN} presents the model-free and model-informed Echo State Network architectures. 
Section \ref{sec:validation} describes the validation strategies. Section \ref{sec:SSV_lor} investigates the robustness of the Single Shot Validation in forecasting chaotic time series.
Section~\ref{sec:Lorenz} analyses the new validation strategies to improve the robustness in forecasting chaotic time series.
Section~\ref{sec:Oscillator} investigates the robustness of the validation strategies in forecasting quasiperiodic time series.
Finally, we summarize the results of this study and discuss future work in the conclusions (section~\ref{sec:Conclusion}).


\section{Echo State Networks}
\label{sec:ESN}

As shown in Fig.~\ref{ESN_scheme}, in the Echo State Network, at any time $t_i$ the input vector, $\textbf{u}_{\mathrm{in}}(t_i) \in \mathbb{R}^{N_u}$, is mapped into the reservoir state, by the input matrix, $\mathbf{W}_{\mathrm{in}} \in \mathbb{R}^{N_r\times N_u}$, where $N_r \gg N_u$. 
The reservoir state, $\textbf{r} \in \mathbb{R}^{N_r}$,  is updated at each time iteration as a function of the current input and its previous value
\begin{equation}
\label{state_step}
        \textbf{r}(t_{i+1}) = \textrm{tanh}\left(\mathbf{W}_{\mathrm{in}}\textbf{u}_{\mathrm{in}}(t_i)+\mathbf{W}\textbf{r}(t_i)\right),
\end{equation}
where $\mathbf{W} \in \mathbb{R}^{N_r\times N_r}$ is the state matrix. The predicted output, $\textbf{u}_{\mathrm{p}}(t_{i+1})\in \mathbb{R}^{N_u}$, is obtained as 
\begin{equation}
\label{output_step}
 \mathbf{u}_{\mathrm{p}}(t_{i+1}) = \hat{\mathbf{r}}(t_{i+1})^T\mathbf{W}_{\mathrm{out}}, \qquad \hat{\mathbf{r}}(t_{i+1}) = \mathbf{g}(\mathbf{r}(t_{i+1}));
\end{equation}
where $\textbf{g}(\cdot)$ is a nonlinear transformation, $\hat{\mathbf{r}} \in \mathbb{R}^{N_{\hat{r}}}$ is the updated reservoir state, and $\mathbf{W}_{\mathrm{out}} \in \mathbb{R}^{N_{\hat{r}}\times N_{u}}$ is the output matrix.
The input matrix, $\mathbf{W}_{\mathrm{in}}$, and state matrix, $\mathbf{W}$, are (pseudo)randomly generated and fixed, while the weights of the output matrix, $\mathbf{W}_{\mathrm{out}}$, are computed by training the network.
In this work, the input matrix, $\mathbf{W}_{\mathrm{in}}$, has only one element different from zero per row, which is sampled from a uniform distribution in $[-\sigma_{\mathrm{in}},\sigma_{\mathrm{in}}]$, where $\sigma_{\mathrm{in}}$ is the input scaling. 
The state matrix, $\textbf{W}$, is an Erdős-Renyi matrix with average sparseness $s$, in which each neuron (each row of $\mathbf{W}$) has on average only $(1-s)N_r$ connections (non-zero elements). 
The non-zero elements are obtained by sampling from a uniform distribution in $[-1,1]$; the entire matrix is then rescaled by a multiplication factor to set the spectral radius, $\rho$. 
The spectral radius is key to enforcing the \emph{echo state property}. 
(In a network with the echo state property, the state loses its dependence on its previous values for sufficiently large times and, therefore, it is uniquely defined by the sequence of inputs.) 
While the echo state property may hold for a wider range of spectral radii~\cite{YILDIZ20121}, the condition $\rho<1$ ensures the echo state property in most situations \cite{lukovsevivcius2012practical}.

The ESN can be run either in open-loop or closed-loop configuration. 
In the open-loop configuration, first, we feed data as the input at each time step to compute and store $\hat{\mathbf{r}}(t_i)$ (\ref{state_step}-\ref{output_step}). 
In the initial transient of this process, the washout interval, we do not compute the output, $\textbf{u}_{\mathrm{p}}(t_i)$. 
The purpose of the washout interval is for the reservoir state to satisfy the echo state property, thereby becoming independent of the arbitrarily chosen initial reservoir state, $\textbf{r}(t_0) = {0}$. 
Secondly, we train the output matrix, $\mathbf{W}_{\mathrm{out}}$, by minimizing the Mean Square Error (MSE) between the outputs, $\textbf{u}_{\mathrm{p}}(t_i)$, and the data, $\textbf{u}_{\mathrm{d}}(t_i)$, over a training set of $N_{\mathrm{tr}}$ points
\begin{equation}
\label{MSE_eq}
    \textrm{MSE} \triangleq \frac{1}{N_{\mathrm{tr}}N_{u}}
    \sum_i^{N_{\mathrm{tr}}} || \textbf{u}_{\mathrm{p}}(t_i) - \textbf{u}_{\mathrm{d}}(t_i)||^2,
 \end{equation}
where $||\cdot||$ is the $L_2$ norm. 
Minimizing \eqref{MSE_eq} is a least-squares minimization problem, which can be solved as a linear system through ridge regression
\begin{equation}
\label{RidgeReg}
    (\mathbf{R}\mathbf{R}^T + \beta \mathbf{I})\mathbf{W}_{\mathrm{out}} = \mathbf{R} \mathbf{U}_{\mathrm{d}}^T,
\end{equation}
\noindent where $\mathbf{R}\in\mathbb{R}^{N_{\hat{r}}\times N_{\mathrm{tr}}}$ and $\mathbf{U}_{\mathrm{d}}\in\mathbb{R}^{N_u\times N_{\mathrm{tr}}}$ are the horizontal concatenation of the updated reservoir states, $\hat{\mathbf{r}}$, and the data, $\textbf{u}_{\mathrm{d}}$, respectively; 
$\mathbf{I}$ is the identity matrix and $\beta$ is the user-defined Tikhonov regularization parameter \cite{tikhonov2013numerical}. 
We solve the linear system through the {\tt linalg.solve} function in {\tt numpy} \cite{harris2020array}. 
In the closed-loop configuration, starting from an initial data point as an input and an initial reservoir state obtained after the washout interval, the output, $\textbf{u}_{\mathrm{p}}$, is fed back to the network as an input for the next time step prediction. 
In doing so, the network is able to autonomously evolve in the future. 
The closed-loop configuration is used during validation and testing.

\subsection{Model-free and model-informed architectures}
\begin{figure}[H]
    \centering
    \subfloat {{\includegraphics[width=0.48\textwidth]{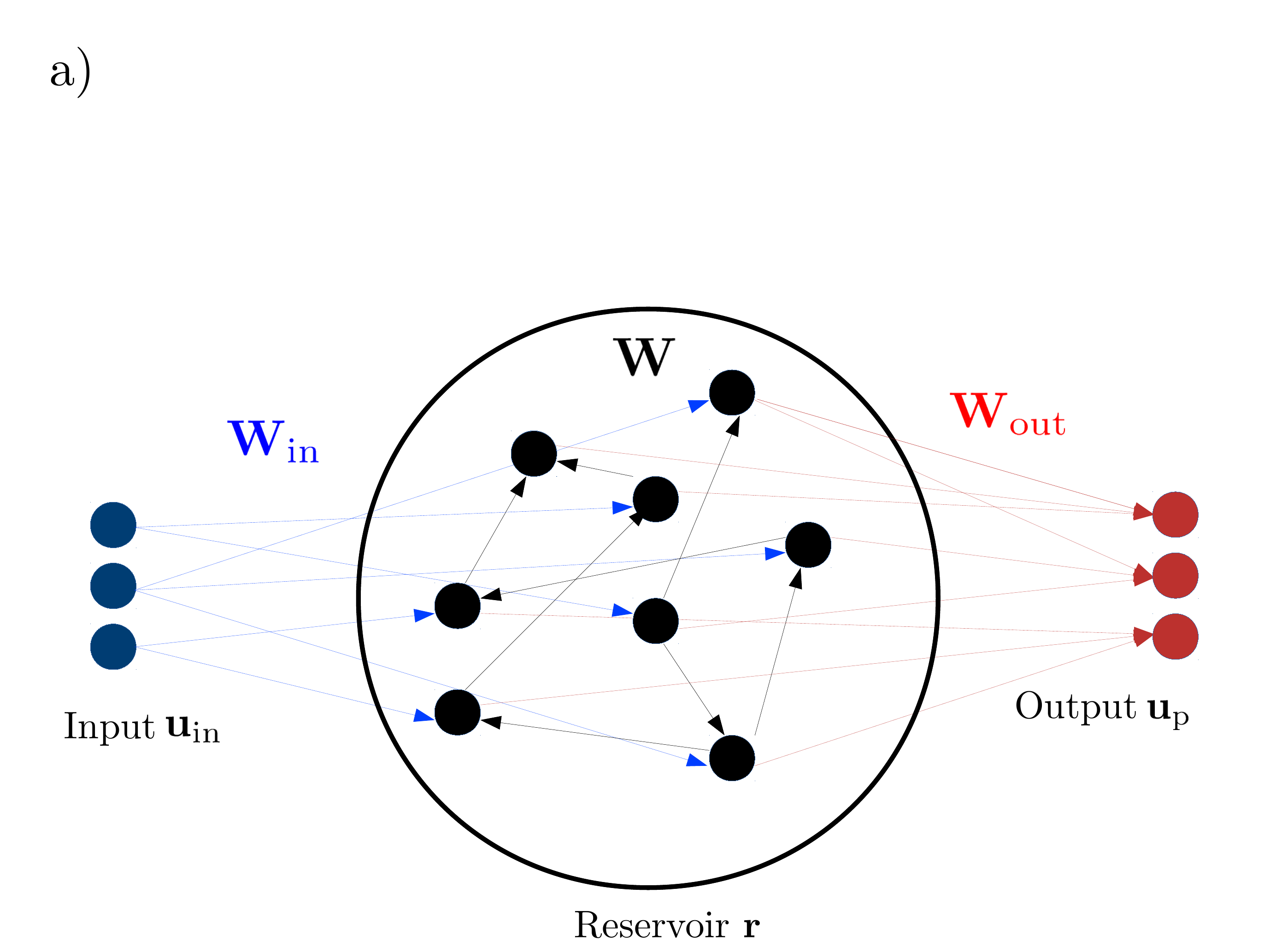} }}%
    \quad
    \subfloat {{\includegraphics[width=0.48\textwidth]{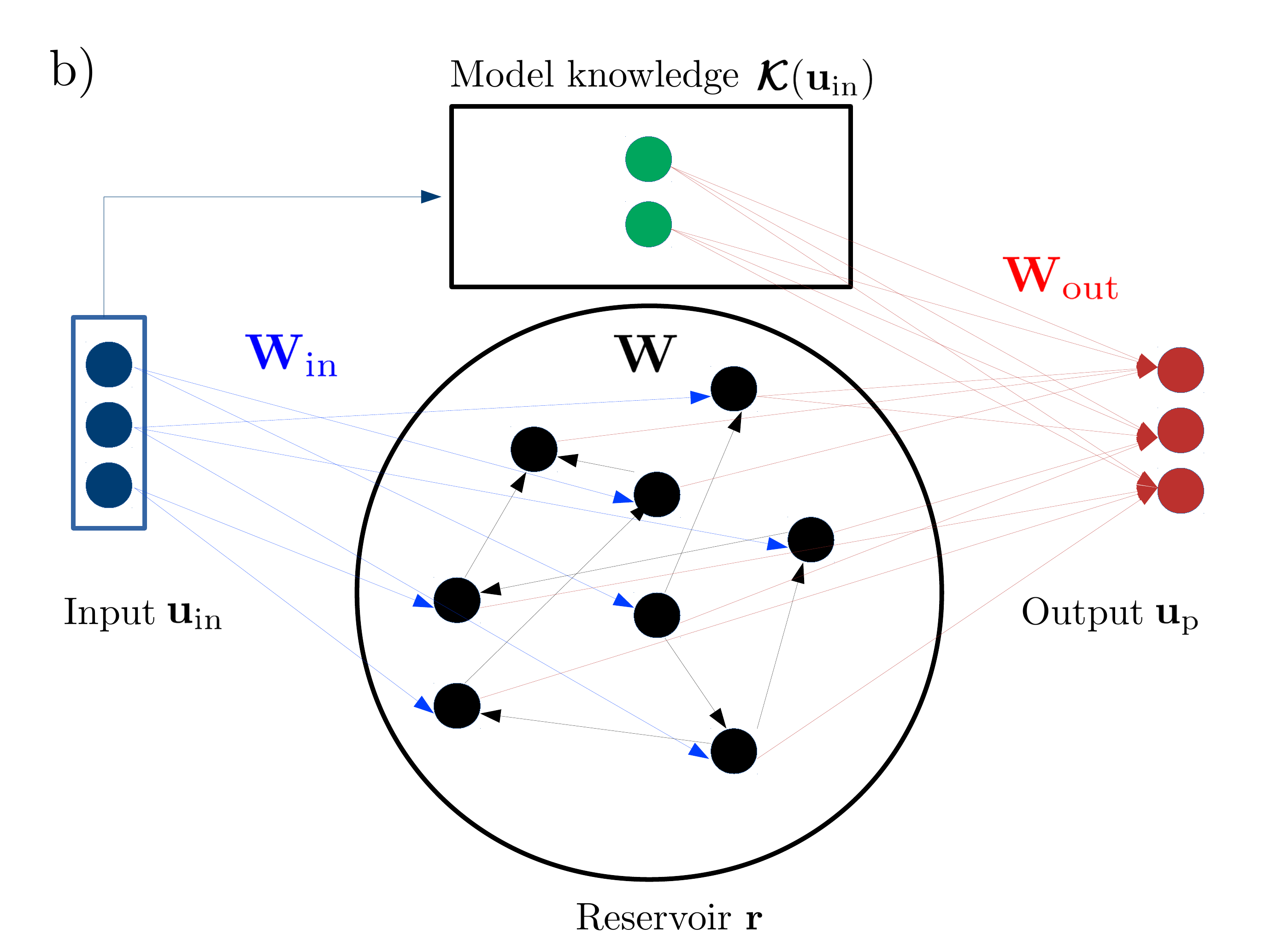} }}%
    \caption{Schematic representation of  (a) model-free and (b) model-informed Echo State Networks (ESNs).}
    \label{ESN_scheme}
\end{figure}

We consider model-free and model-informed architectures (Fig. \ref{ESN_scheme}). 
The basic model-free ESN is obtained by setting $\textbf{g}(\textbf{r}) = \textbf{r}$. 
This architecture, however, generates symmetric solutions in the closed-loop configuration \cite{lu2017reservoir, huhn2020learning}, which can cause the predicted trajectory to stray away from the actual attractor towards a symmetric attractor, which is not a solution of the dynamical system (but it is a solution of the network). 
To break the symmetry, we add biases in the input and output layers 
\begin{equation}
    \textbf{r}_{i+1} = \textrm{tanh}\left(\mathbf{W}_{\mathrm{in}}[\textbf{u}_{\mathrm{in}};b_{\mathrm{in}}]+\mathbf{W}\textbf{r}_i\right), \qquad \hat{\mathbf{r}}_i = [\textbf{r}_{i+1};1], \qquad \textbf{u}_{\mathrm{p}} = \hat{\mathbf{r}}_i^T\mathbf{W}_{\mathrm{out}} \label{arch_bias}.
\end{equation}
\noindent where $[\, \cdot \,; \, \cdot \,]$ indicates vertical concatenation, $b_{\mathrm{in}}$ is the scalar input bias and $\mathbf{W}_{\mathrm{in}} \in \mathbb{R}^{N_r\times(N_u+1)}$. 
In the model-informed ESN, also known as hybrid as proposed by~\cite{pathak2018hybrid}, information about the governing equations (model knowledge) is embedded into the model through a function of the input, $\pmb{\mathcal{K}}(\textbf{u}_{\mathrm{in}})$, which, for example, may be a reduced order model that provides information about the output at the next time step as
\begin{equation}
   \hat{\mathbf{r}}_i = [\textbf{r}_{i+1};1;\pmb{\mathcal{K}} (\textbf{u}_{\mathrm{in}})]. 
  \label{eq:Hyb_ESN}
\end{equation}
In this work, we use $\pmb{\mathcal{K}}(\textbf{u}_{\mathrm{in}})$ only to update the reservoir state \cite{wikner2020combining}, in order to use the same input matrix, $\textbf{W}_{\mathrm{in}}$, and state matrix, $\textbf{W}$, of the model-free architecture.
 This allows us to directly compare the performances of the model-free and model-informed architectures. 

\section{Validation}
\label{sec:validation}

The purpose of the validation is to determine the hyperparameters by minimizing an error. 
We make a distinction between the hyperparameters (i) that require re-initialization of $\mathbf{W}_{\mathrm{in}}$ and $\mathbf{W}$, 
and (ii) that do not require re-initialization. 
The size of the reservoir, $N_r$, and sparseness, $s$, require re-initialization, whereas the input scaling, $\sigma_{\mathrm{in}}$, the spectral radius, $\rho$, the Tikhonov parameter, $\beta$, and the input bias $b_{\mathrm{in}}$, do not. 
The fundamental difference between (i) and (ii) is that the random component of the re-initialization of $\mathbf{W}_{\mathrm{in}}$ and $\mathbf{W}$ makes the objective function to be minimized random, which significantly increases the complexity of the optimization. 
In this study, we minimize the error with respect to the input scaling, $\sigma_{\mathrm{in}}$, and spectral radius, $\rho$, which are  key hyperparameters for the performance of the network \cite{jiang2019model, lukovsevivcius2012practical}.  
For convenience, we  rewrite the reservoir state equation (\ref{state_step}) as 
\begin{equation}
 \textbf{r}_{i+1} = \textrm{tanh}(\sigma_{\mathrm{in}} \mathbf{\hat{W}}_{\mathrm{in}}[\textbf{u}_{\mathrm{in}};b_{\mathrm{in}}]+\rho \mathbf{\hat{W}}\textbf{r}_i),
 \end{equation}
where the non-zero elements of $\mathbf{\hat{W}}_{\mathrm{in}}$ are sampled from the uniform distribution in [-1,1] and $\mathbf{\hat{W}}$ has been scaled to have a unitary spectral radius. 

\subsection{Performance metrics}
We determine the hyperparameters by minimizing the Mean Squared Error \eqref{MSE_eq} in the validation interval of fixed length. 
The networks are tested on multiple starting points along the attractor by using both the Mean Squared Error and Prediction Horizon (PH), the latter of which is defined as the time interval during which the normalized error is smaller than a user-defined threshold $k$ \cite{doan2019physics, pathak2018hybrid}
\begin{equation}
\label{PH}
    \frac{|| \textbf{u}_{\mathrm{p}}(t_i) - \textbf{u}_{\mathrm{d}}(t_i)||} {\sqrt{\frac{1}{N_{\mathrm{PH}}}\sum_j^{N_{\mathrm{PH}}}||\textbf{u}_{\mathrm{d}}(t_j)||^2}}
    < k,
\end{equation}
where $N_{\mathrm{PH}}$ are the number of timesteps in the Prediction Horizon. The Mean Squared Error and Prediction Horizon for the same starting point in the attractor are strictly correlated (\ref{A:Val_Fig}). 
We use the Mean Squared Error to partition the dataset in intervals of fixed length during validation, while we use the Prediction Horizon in the test set because it is the most physical quantity when assessing the time-accurate prediction of chaotic systems~\cite[e.g.,][]{pathak2018model,DOAN2020}.

\subsection{Strategies}

\label{sec:valstrat}

\begin{figure}[H]
    \centering
    \includegraphics[width=1.0\textwidth]{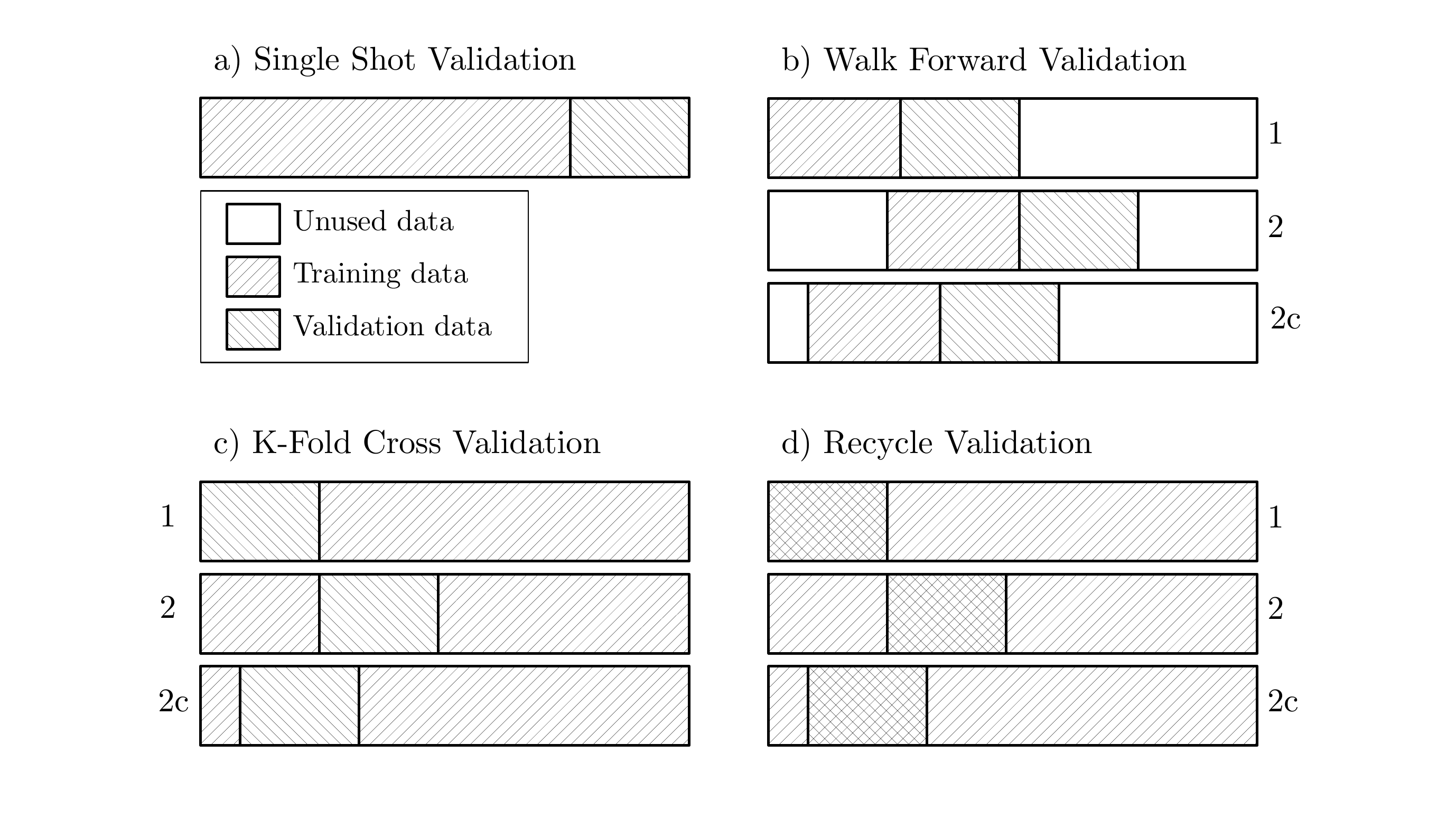}
    \caption{Partition of the data in the different validation strategies. In (b-d),  bar 1 shows the first fold,   bar 2 shows the second fold, and  bar 2c shows the second fold in the chaotic version (shifted by one Lyapunov time).}
    \label{Val_Strat}
\end{figure}

The most common validation strategy for ESNs is the Single Shot Validation (SSV), which splits the available data in a training set and a single subsequent validation set (Fig.~\ref{Val_Strat}a). 
The time interval of the validation set, during which the hyperparameters are tuned, is small and represents only a fraction of the attractor. 
In nonlinear time series prediction, the choice of the validation strategy has to take into account (i) the intervals we are interested in predicting and (ii) the nature of the signal we are trying to reproduce. 
Here, we are interested in predicting multiple intervals as the trajectory spans the attractor, rather than a specific interval starting from a specific initial condition. 
Moreover, an ergodic trajectory of the 
attractor has no underlying  time-varying  statistics, e.g there is no time-dependency of the mean of the signal, hence trajectories return indefinitely in nearby regions of the attractor. 
This means that the intervals we are interested in predicting are potentially similar to any interval of the trajectory that constitutes our dataset, regardless of the interval position in time within the dataset. 
For this reason, as shown in section~\ref{sec:SSV_lor}, the Single Shot Validation strategies should not be employed in chaotic time series.

These observations lead us to use validation strategies based on multiple validation intervals, which may precede the training set, such as the Walk Forward Validation (WFV) and the K-Fold cross Validation (KFV).
We also propose an ad-hoc robust validation strategy---the Recycle Validation (RV). 
The objective of these strategies is to tune the hyperparameters over an effectively larger portion of the trajectory, by minimizing the average of the objective function (error) over multiple validation intervals. 
The regular version of these strategies consists of creating subsequent folds by moving forward in time the validation set by its own length.
Additionally, we propose the chaotic version, in which we move the fold forward in time by one Lyapunov Time (LT) (Fig. \ref{Val_Strat}). 
The Lyapunov Time is a key time scale in chaotic dynamical systems, which is defined as the inverse of the leading Lyapunov exponent $\Lambda$ of the system, which, in turn, is the exponential rate at which infinitesimally close trajectories, $\delta\mathbf{q}(0)$ diverge~\cite[e.g.,][]{Boffetta2002}
\begin{align}
\lvert\lvert\delta\mathbf{q}(t) \lvert\lvert\sim \lvert\lvert\delta\mathbf{q}(0)\lvert\lvert \exp(\Lambda t)\quad\quad t\rightarrow\infty, \quad  \lvert\lvert\delta\mathbf{q}(0)\lvert\lvert\rightarrow 0.
\end{align}

\textbf{Walk Forward Validation.} In the Walk Forward Validation (WFV) (Fig. \ref{Val_Strat}b), we partition the available data in multiple splits, while maintaining sequentiality of the data. 
From a starting dataset of length $n$, the first $m$ points ($m<n$) are taken as the first fold, with $N_t$ points for training and $v$ points for validation ($v+N_t=m$). 
These quantities must respect $(n-m) = (k_1-1) v; \; k_1 \in \mathbb{N}$.
The remaining $(k_1-1)$ folds are generated by moving the training plus validation set forward in time by a number of points $v$. 
This way, the original dataset is partitioned in $k_1$ folds and the hyperparameters are selected to minimize the average MSE over the folds. \\ 

\textbf{K-Fold cross Validation.} Although the K-Fold cross Validation (KFV) (Fig. \ref{Val_Strat}c) is a common strategy in regression and classification, it is not commonly used in time series prediction because the validation and training intervals are not sequential to each other. 
This strategy partitions the available data in $k_2$ splits.  
Over the entire dataset of length $n$, after an initial $bv$ points, with $0\leq b<1$, needed to have an integer number of splits, the remaining $n-bv$ points are used as $k_2$ validation intervals, each of of length $v$. 
For each validation interval we define a different fold, in which we use all the remaining data points for training. 
We determine the hyperparameters by minimizing the average of the MSE between the folds. \\ 

\textbf{Recycle  Validation.} We propose a the Recycle Validation (RV) (Fig. \ref{Val_Strat}d), which exploits the information obtained by both open-loop and closed-loop configurations. Because the network works in two different configurations, it can obtain additional information when validating on data already used in training. 
To do so, first, we train the output weights using the entire dataset of $n$ points.
Second, we validate the network on $k_2$ splits of length $v$ from data that has already been used to train the output weights. 
Each split is imposed by moving forward in time the previous validation interval by $v$ points. 
After an initial $bv$ points, with $0\leq b<1$, needed to have an integer number of splits, the remaining $n-bv$ points are used as $k_2$ validation intervals. 
We determine the hyperparameters by computing the average of the MSE between the splits. 
This strategy has four main advantages.
First, it can be used in small datasets, where the partition of the dataset in separate training and validation sets may cause the other strategies to perform poorly. In small datasets, the validation intervals represent a larger percentage of the dataset since each validation interval needs to be multiple Lyapunov Times to capture the divergence of chaotic trajectories. Therefore, the training set becomes substantially smaller than the dataset and the output matrix used during validation differs substantially from the output matrix of the whole dataset. This results in a poor selection of hyperparameters.
Second, for a given dataset, we maximize the number of validation splits, using the same validation intervals of the K-Fold cross Validation. 
Third, we tune the hyperparameters using the same output matrix, $\mathbf{W}_{\mathrm{out}}$, that we use in the test set. 
Fourth, it has lower computational cost than the K-Fold cross Validation because it does not require retraining the output matrix for the different folds. which makes it computationally cheaper (\ref{A_CompTime}). \\ 

\textbf{Chaotic version.} The chaotic version consists of shifting the validation intervals forward in time, not by their own length, but by one Lyapunov Time when constructing the next fold. 
In doing so, different splits will overlap, but, since the trajectory related to the split that started one Lyapunov Time (LT) earlier has strayed away from the attractor on average by $e^{\Lambda\times 1\mathrm{LT}}=e$, the two intervals contain different information. 
The purpose of this version is to further increase the number of intervals on which the network is validated. 
The regular and  chaotic versions for each validation strategy are shown in frames (b-d) in Fig. \ref{Val_Strat} in bars 2 and 2c, respectively.
The \emph{chaotic versions} of the Walk Forward Validation, the K-fold cross Validation and the Recycle Validation are denoted by the subscript $c$. 

\subsection{Grid search and Bayesian optimization}

\label{sec:BO}

To find the minimum of the Mean Squared Error \eqref{MSE_eq} of the validation set in the hyperparameter space, we use Bayesian Optimization (BO), which is compared to Grid Search (GS). 
Bayesian Optimization has been shown to outperform other state-of-the-art optimization methods when the number of evaluations of an expensive objective function is limited \cite{brochu2010tutorial, snoek2012practical}. 
It is a global search method, which is able to incorporate prior knowledge about the objective function and to use information from the entire search space. 
It treats the objective function as a black box, therefore, it does not require gradient information. 
Starting from an initial $N_{st}$ evaluations of the objective function, BO performs a Gaussian Process (GP) regression \cite{rasmussen2003gaussian} to reconstruct the function in the search space, using function evaluations as data. 
Once the GP fitting is available, we select the new point at which to evaluate the objective function so that the new point maximizes the acquisition function. 
The acquisition function is evaluated on the mean and standard deviation of the GP reconstruction. 
After the objective function is evaluated at a new point, the enlarged data set, comprising of the new point, is used to perform another GP regression, select a new point and so on and forth. 
In this work, we use the gp-hedge Bayesian Optimization algorithm  implemented in {\tt scikit-optimize} library in Python \cite{hoffman2011portfolio, 2020SciPy-NMeth}. 
The details of the formulation are explained in \ref{A_BO} and the Supplementary Material (S.2).

\section{Robustness of the Single Shot Validation}
\label{sec:SSV_lor}

The first testcase we investigate is the Lorenz system \cite{lorenz1963deterministic}, which is a reduced-order model of Rayleigh–Bénard convection 
\begin{align}
    \dot{x} &= \sigma_L(y-x) \nonumber \\
    \dot{y} &= x(\rho_L - z) - y \nonumber \\
    \dot{z} &= xy - \beta_L z,
\end{align}

\noindent where $[\sigma_L,\beta_L,\rho_L] = [10,8/3,28]$ is selected to generate chaotic solutions~\cite[e.g.,][]{lorenz1963deterministic}. 
The system is integrated with a forward Euler scheme with step $dt=0.009$ LT.
The Lyapunov Time is $\mathrm{LT}=\Lambda^{-1}\approx 1.1$ \cite{viswanath1998lyapunov}. \\

\begin{figure}[H]
    \centering
    \includegraphics[width=1.0\textwidth]{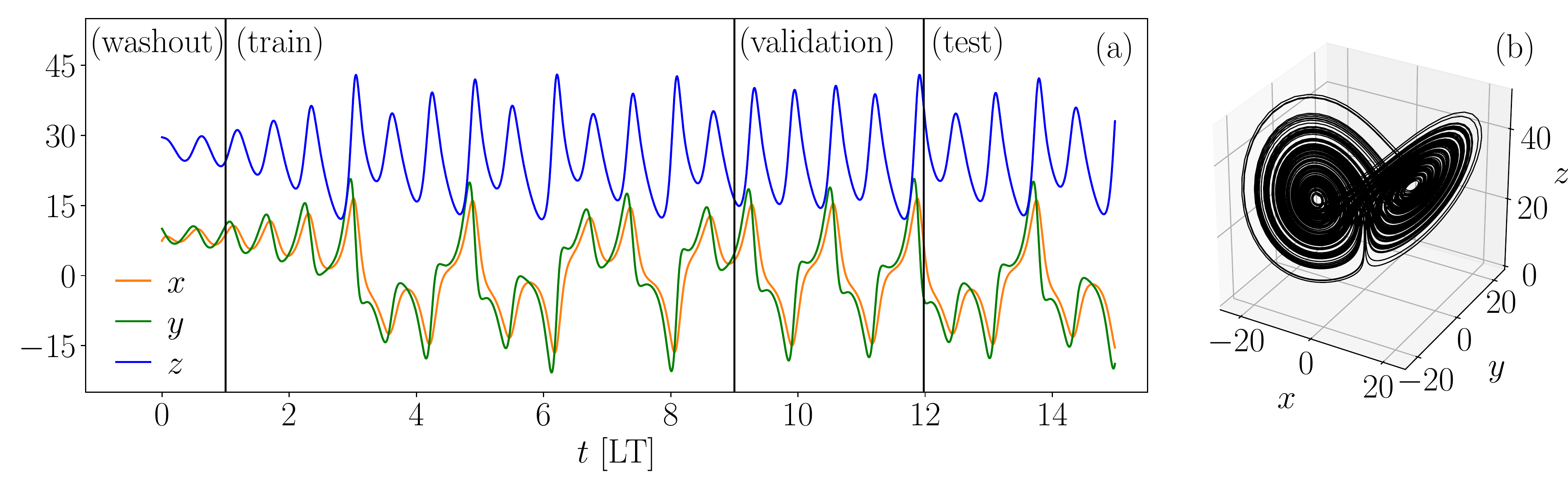}
    \caption{ Solution of the Lorenz system. (a) Time series, and (b) phase plot for a longer time window. Time is expressed in Lyapunov time (LT) units. 
    }
    \label{Lorenz Time Series}
\end{figure}

We analyse the performance of the  used Single Shot Validation (SSV), which is employed for training (1LT to 9LTs), validation (9LTs to 12 LTs), and testing (12 LTs to 15LTs), as shown in Fig. \ref{Lorenz Time Series}. 
The input, $\mathbf{u}_{\mathrm{in}}$, is normalized by its maximum variation. 
(This is done because we are using a single scalar quantity $\sigma_{\mathrm{in}}$ to scale all the components of the input.) 
The network has a fixed number of neurons, $N_r=100$, sparseness, $s=97\%$, Tikhonov parameter, $\beta_t = 10^{-11}$ and input bias, $b_{\mathrm{in}}=1$ \cite{lukovsevivcius2012practical}. 
The bias, $b_{\mathrm{in}}$, is set for it to have the same order of magnitude of the normalized input. 
The input scaling, $\sigma_{\mathrm{in}}$, and spectral radius, $\rho$, are tuned during validation in the range $[0.5,5]\times[0.1,1]$ to minimize the $\log_{10}$(MSE). 
The range of the spectral radius, $\rho$, is selected for the network to respect the echo state property, whereas the range of the input scaling, $\sigma_{\mathrm{in}}$, is selected to normalize the inputs. 
The optimization is performed with 
(i)  Grid Search (GS) consisting of 7$\times$7 points, 
and (ii) Bayesian Optimization (BO) consisting of 5$\times$5 starting points and 24 points acquired by the gp-hedge algorithm (\ref{A_BO}). 
The two optimization schemes are applied to an ensemble of $N_{\mathrm{ens}}=50$ networks, which differ by the random initialization of the input matrix, $\mathbf{W}_{\mathrm{in}}$, and the state matrix, $\mathbf{W}$. $N_{\mathrm{ens}}$ is selected after a test on statistical convergence (\ref{A:Convergence}).  

\begin{figure}[H]
    \centering
    \includegraphics[width=1.\textwidth]{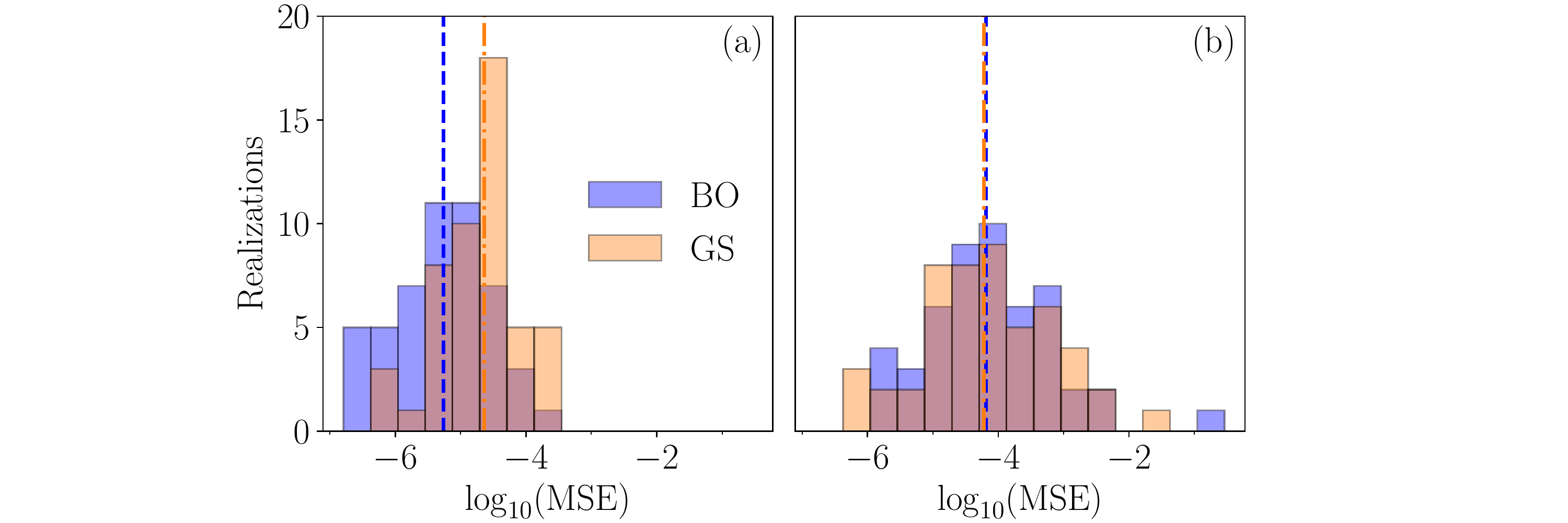}
    \caption{Performance of the optimal hyperparameters computed by Grid Search (GS) and  Bayesian Optimization (BO) in (a) validation and (b) test sets.
    Vertical lines indicate the median of  Grid Search (dash-dotted) and  Bayesian Optimization (dashed).
    The medians are $[5.4,23.0]\times10^{-6}$ in the validation set and $[64.8,60.5]\times10^{-6}$ in the test set for BO and GS, respectively.
    }
    \label{ValTest_Set_MSE}
\end{figure}

\begin{figure}[H]
    \centering
    \includegraphics[width=1.\textwidth]{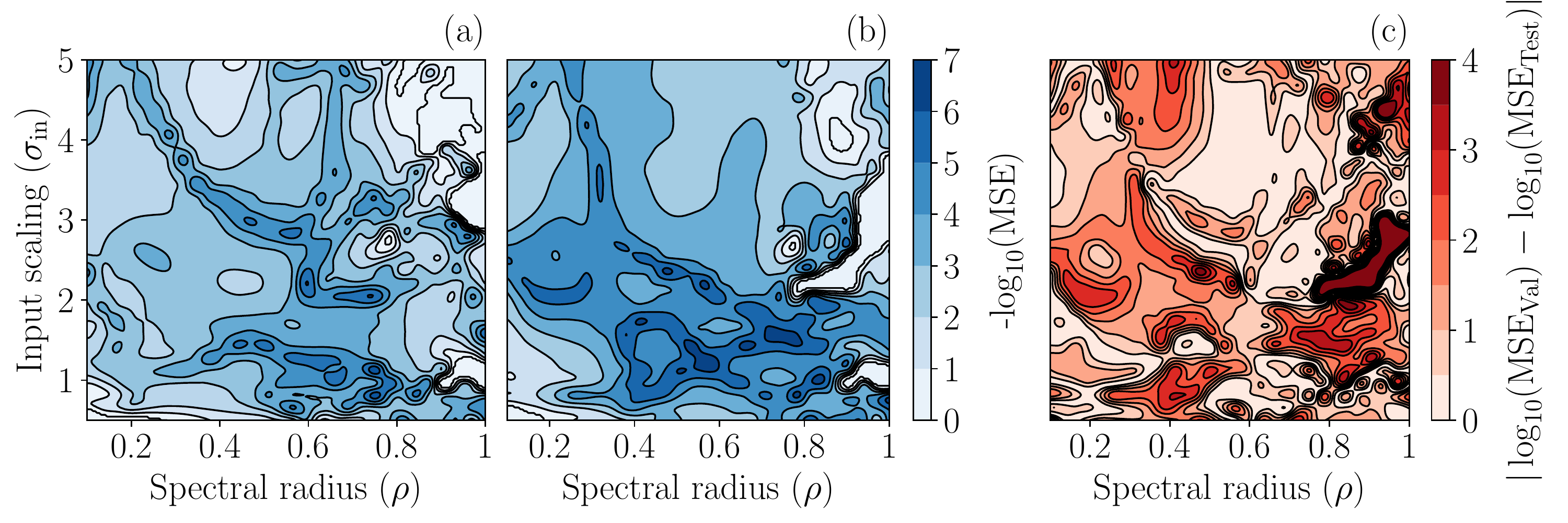}
    \caption{Mean of the Gaussian Process reconstruction in the (a) validation and (b) test sets for a representative network in the ensemble. 
    Frame (c) shows the difference between the two sets. 
    The $\mathrm{MSE}$ is saturated to be $\leq 1$ in (a,b), whereas the error is saturated to be $\leq10^4$ in (c). 
The reconstruction is performed on a grid of 30$\times$30  evaluations of $\log_{10}(\mathrm{MSE})$.
    For the same hyperparameters, the MSE can differ by orders of magnitude between the validation and test sets. 
    }
    \label{MSE_Surfaces}
\end{figure}

Figure~\ref{ValTest_Set_MSE} shows the performance of the optimal hyperparameters computed by Grid Search and Bayesian Optimization for the ensemble members.  
First, we analyse the performance in validation (panel (a)).
As shown by the medians reported in the caption, Bayesian optimization markedly outperforms Grid Search. 
Second, we analyse the performance in the test set (panel (b)). 
The performance of each network is assessed by computing the MSE in the test set for the hyperparameters found in the validation set. 
For this, the output matrix, $\mathbf{W}_{\mathrm{out}}$, of the test set is obtained by retraining over both the training and validation sets. 
As shown by the medians, the overall performance of the networks and the benefit of using Bayesian Optimization are markedly reduced.
This is a signature of chaos, whose unpredictability results in a weak correlation between validation and test sets.
This is further verified by computing the mean of the Gaussian process reconstruction from a 30$\times$30 grid of  $\log_{10}(\mathrm{MSE})$ for a representative network of the ensemble  (Fig.~\ref{MSE_Surfaces}). 
%
The performance of the optimal hyperparameters of the validation set can deteriorate by four, or more, orders of magnitude in the test set (panel (c)). 

\begin{figure}[H]
    \centering
    \includegraphics[width=1.0\textwidth]{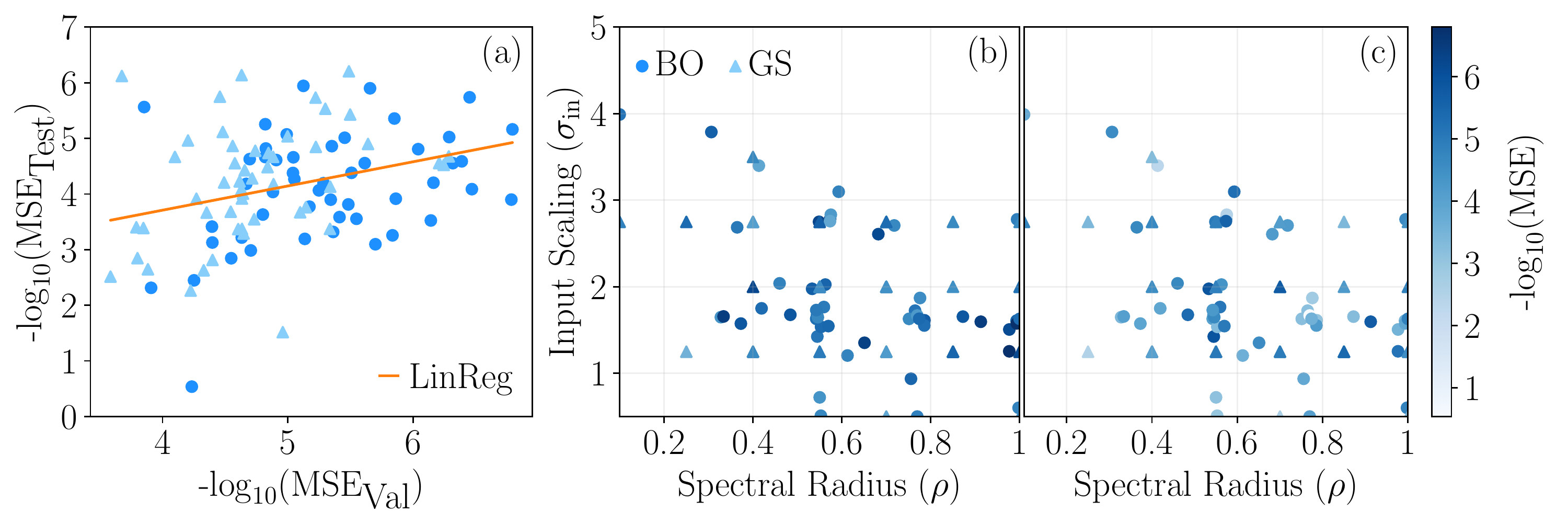}
    \caption{(a) Linear regression (LinReg) and scatter plot of the MSE of the optimal hyperparameters obtained from Bayesian Optimization (BO) and Grid Search (GS) for each network. 
    Optimal hyperparameters for each network and corresponding MSE in (b) validation and (c) test sets. 
    For different networks the optimal hyperparameters, and their performance, vary significantly.
    }
    \label{Lack_rob}
\end{figure}

To  assess quantitatively the correlation of the optimal hyperparameters' performance between the validation and test sets, we use the Spearman coefficient \cite{spearman1904} 
\begin{align}
     \tilde{r}_S(\textbf{x},\textbf{y}) & = \frac{\sum_i(z(x)_{i}-N_{\mathrm{ens}})(z(y)_{i}-N_{\mathrm{ens}})}{\sqrt{\sum_i(z(x)_{i}-N_{\mathrm{ens}})^2}\sqrt{\sum_i(z(y)_{i}-N_{\mathrm{ens}})^2}} , \nonumber\\ 
     \mathbf{x}  &=    
     \left[
     \begin{array}{c} 
           \textbf{m}_{\mathrm{Val}}^{\mathrm{(BO)}}\\
           \textbf{m}_{\mathrm{Val}}^{\mathrm{(GS)}} \\
        \end{array}
        \right], \quad\quad 
             \mathbf{y}  =    
     \left[
     \begin{array}{c} 
           \textbf{m}_{\mathrm{Test}}^{\mathrm{(BO)}}\\
           \textbf{m}_{\mathrm{Test}}^{\mathrm{(GS)}} \\
        \end{array}
        \right],
    \label{r-tilde}
\end{align}
where $\mathbf{z(x)}$ is the ranking function; $\mathbf{m} \in \mathbb{R}^{N_{\mathrm{ens}}}$ contains the MSE for the optimal hyperparameters in validation (subscript Val), or test (subscript Test) obtained by Bayesian Optimazation (superscript BO), or Grid Search (superscript GS). 
$\tilde{r}_{S}$ quantifies the correlation between the MSE of the optimal hyperparameters obtained during validation and the MSE for the same hyperparameters in the test set over the ensemble. 
The values $\tilde{r}_S=\{-1,0,1\}$ indicate anticorrelation, no correlation and correlation, respectively.
Figure~\ref{Lack_rob} shows the correlation analysis. 
The scatter plot for $\mathbf{x}$ and $\mathbf{y}$ (panel (a))
shows that the MSE of the optimal hyperparameters in the validation and test sets are weakly correlated with $\tilde{r}_S=0.32$,  independently on whether they are computed with Bayesian Optimization or Grid Search. 
Panels (b,c) show the values of the optimal hyperparameters, which vary substantially from one network realization to another. 

\subsection{Remarks}
First, because the MSE and optimal hyperparameters vary significantly in different network realizations, we advise performing optimization separately for each network to increase the accuracy (as further verified in \ref{sec: A_fix}). 
Second, hyperparameters that are optimal in the validation set may have a poor performance in the test test, which may greatly reduce the benefit of using  Bayesian hyperparameter optimization. 
This highlights a fundamental challenge in learning chaotic solutions, in which validation and test sets may be topologically different portions of the attractor. 
We, thus, advise that the Single Shot Validation not be used in the validation of Echo State Networks in chaotic attractors. 
Robust validation strategies (section \ref{sec:valstrat}) are next analysed.   

\section{Validation for chaotic solutions}
\label{sec:Lorenz}
\subsection{Hyperparameter optimization}

\label{sec:Lor_Val_Strat}
We compare different validation strategies on the ensemble of $N_{ens}=50$ networks in a ``short'' dataset (12 LTs) and a ``long'' dataset (24 LTs). 
The long dataset is obtained by the integration of the time series in Fig. \ref{Lorenz Time Series}. 
In addition to the short dataset, we analyse the long dataset for two reasons. 
First, we wish to test validation strategies that require larger datasets to fully perform, such as the Walk Forward Validation. 
Second, we wish to investigate how the robustness is affected by the size of the dataset. 
We use the Single Shot Validation (SSV), Walk Forward Validation (WFV), K-Fold Validation (KFV), Recycle Validation (RV), and corresponding chaotic versions (subscript $c$). 
The long dataset allows us to define an additional chaotic Walk Forward Validation (WFV$_{\mathrm{c}}$) denoted by the superscript $*$ as detailed in the Supplementary Material (S.1). \\

The test set has $N_t=100$ starting points on the attractor to sample different regions of the solutions (more details in~\ref{A:Convergence}).  
The Prediction Horizon is globally quantified as an arithmetic mean, $\overline{\mathrm{PH}}_{\mathrm{test}}$, with threshold $k=0.2$; whereas the Mean Squared Error is globally quantified as a geometric mean, $\overline{\mathrm{MSE}}_{\mathrm{Test}}$, in intervals of 3LTs.

\subsubsection{Model-free ESN}

\begin{figure}[H]
    \centering
    \includegraphics[width=1.\textwidth]{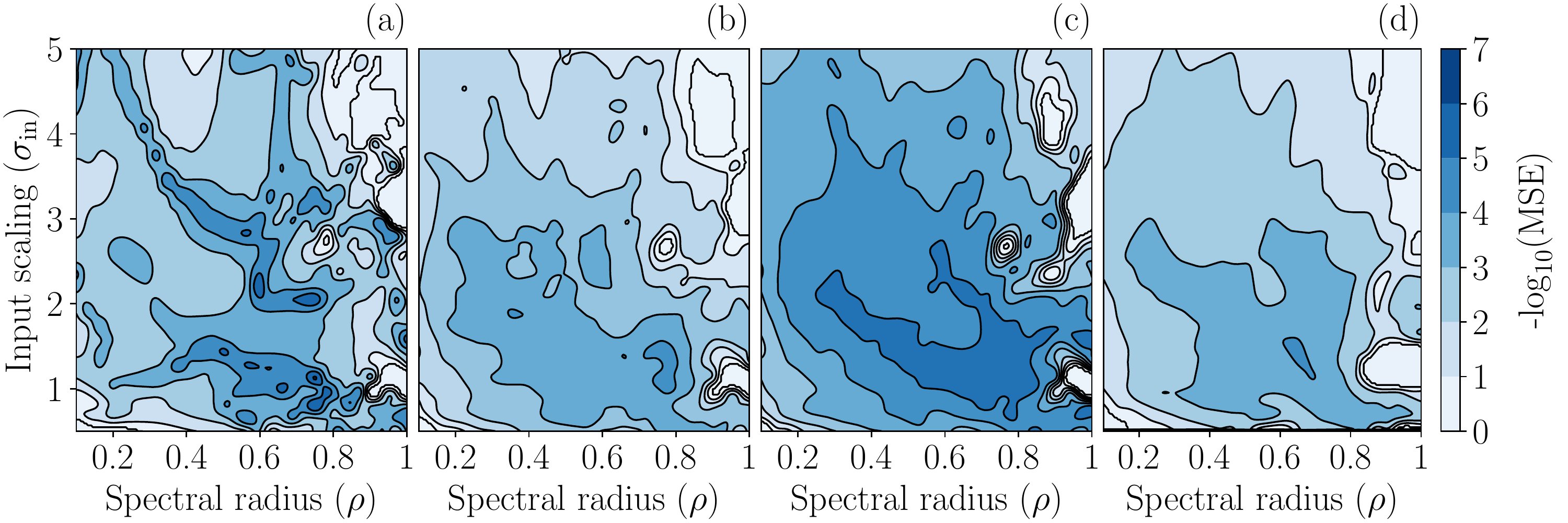}
    \caption{Mean of the Gaussian Process reconstruction for the short dataset for a representative network of the ensemble. Validation set for (a) Single Shot Validation (SSV), (b) chaotic K-Fold Validation (KFV$_{\mathrm{c}}$), and (c) chaotic Recycle Validation (RV$_{\mathrm{c}}$); and test set (d). The $\mathrm{MSE}$ is saturated to be $\leq 1$.  The Gaussian Process is based on a grid of 30$\times$30 data points. 
    }
    \label{Strat_MSE_Surfaces}
\end{figure}

Figure~\ref{Strat_MSE_Surfaces} shows the mean of the Gaussian Process reconstruction of $\log_{10}(\mathrm{MSE})$ in the hyperparameter space for a representative network of the ensemble.
Panels (a,b,c) show the performance of three validation strategies in the validation set, whereas panel (d) shows the performance of the network in the test set. 
Because the error in (b,c) is similar to the error in (d), and the error in (a) differs from (d), we conclude that  in the test set 
the hyperparameters computed through KFV$_{\mathrm{c}}$ and RV$_{\mathrm{c}}$ perform well, but the hyperparameters computed through SSV perform poorly.  

\begin{figure}[H]
    \centering
    \includegraphics[width=1.0\textwidth]{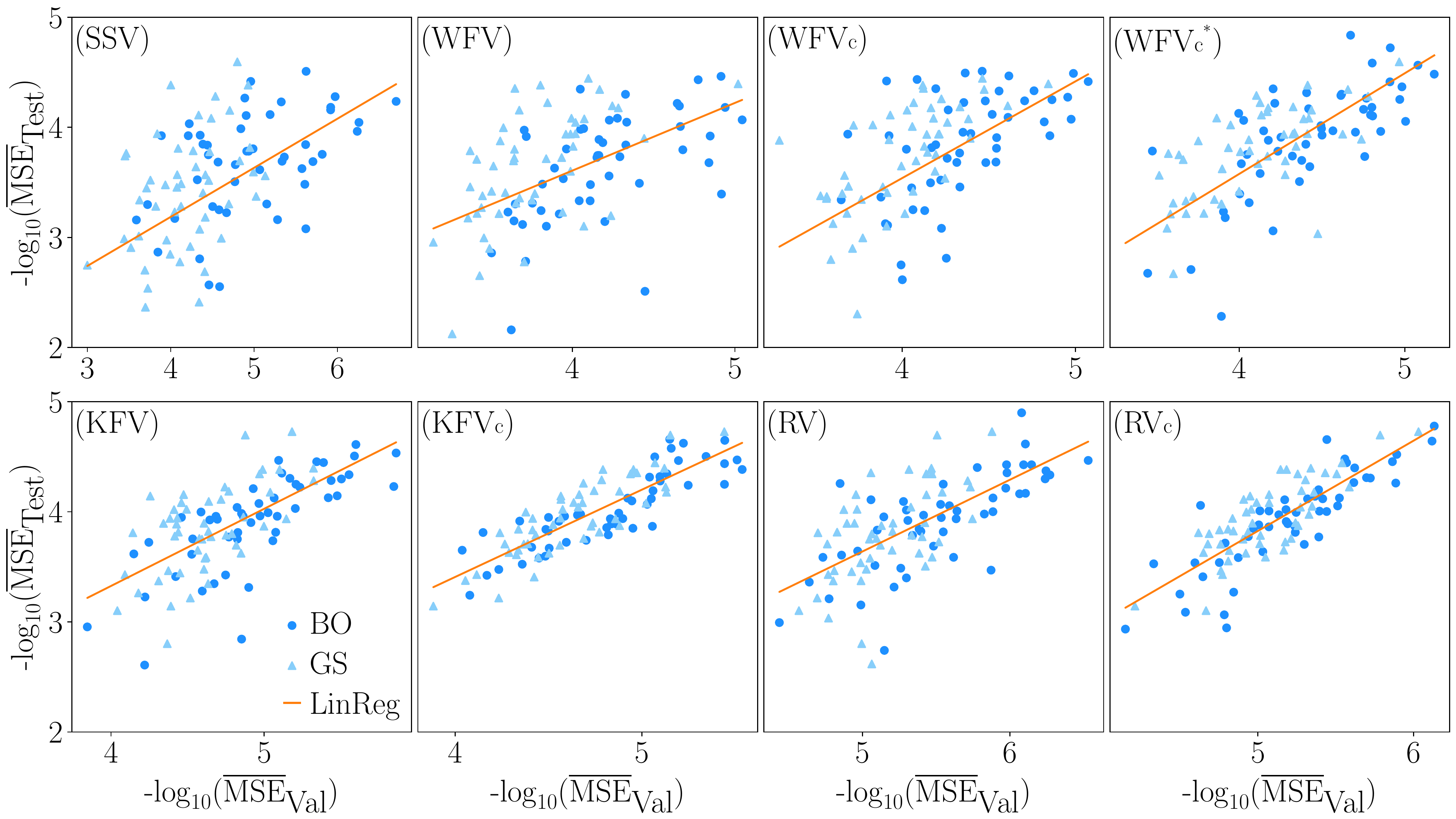}
    \caption{Linear regression (LinReg) and scatter plot of the MSE of the optimal hyperparameters obtained from Bayesian Optimization (BO) and Grid Search (GS) for each network. 
    Single Shot Validation (SSV), Walk Forward Validation (WFV), K-Fold Validation (KFV), Recycle Validation (RV), and their chaotic versions (subscript  $c$). Long dataset. 
    }
    \label{Lorenz_Correlations}
\end{figure}

A correlation analysis is shown in Tab. \ref{tab:24LT_12LT} with the Spearman correlation coefficients, $\Tilde{r}_s$ (\ref{r-tilde}) (short and long datasets); 
and  Fig. \ref{Lorenz_Correlations} with scatter plots of the optimal hyperparameters' performance (long dataset, for brevity). 
The Single Shot Validation has the lowest correlation among all the validation strategies in both datasets. 
The chaotic versions of the validation strategies correlate better than the corresponding regular versions. 
In particular, the chaotic K-Fold Validation and the chaotic Recycle Validation have the highest correlations.
In general, increasing the size of the dataset increases the correlation, but the Single Shot Validation in the long dataset has a lower correlation than the K-Fold Validations and the Recycle Validations in the short dataset. This further demonstrates the poor robustness of the Single Shot Validation.  Last, but not least, the Recycle Validation is computationally cheaper than the K-Fold Validation because the output matrix is the same for the different folds (more analysis on the computational time can be found in~\ref{A_CompTime}).

\begin{table}[H]
\centering
    \caption{Spearman  coefficients between validation and test sets. Bold text indicates the highest correlation in the dataset. }
\begin{tabular}{c c c c c c c c c c}
$\Tilde{r}_S$ & $\,$ &  SSV & WFV &  WFV$_{\mathrm{c}}$ &  WFV$_{\mathrm{c}}^*$ & KFV & KFV$_{\mathrm{c}}$ & RV & RV$_{\mathrm{c}}$ \\
\toprule
Short dataset  (12LTs) & & 0.31 &0.31 & 0.50  & - & 0.60 & \textbf{0.65} & 0.59 & 0.62 \\
\midrule 
Long dataset (24LTs) & & 0.49 & 0.51 & 0.61 & 0.70 & 0.70 & \textbf{0.85} & 0.67 & 0.81 \\
\toprule
\end{tabular}
\label{tab:24LT_12LT}
\end{table}

\begin{figure}[H]
    \centering
    \includegraphics[width=1.0\textwidth]{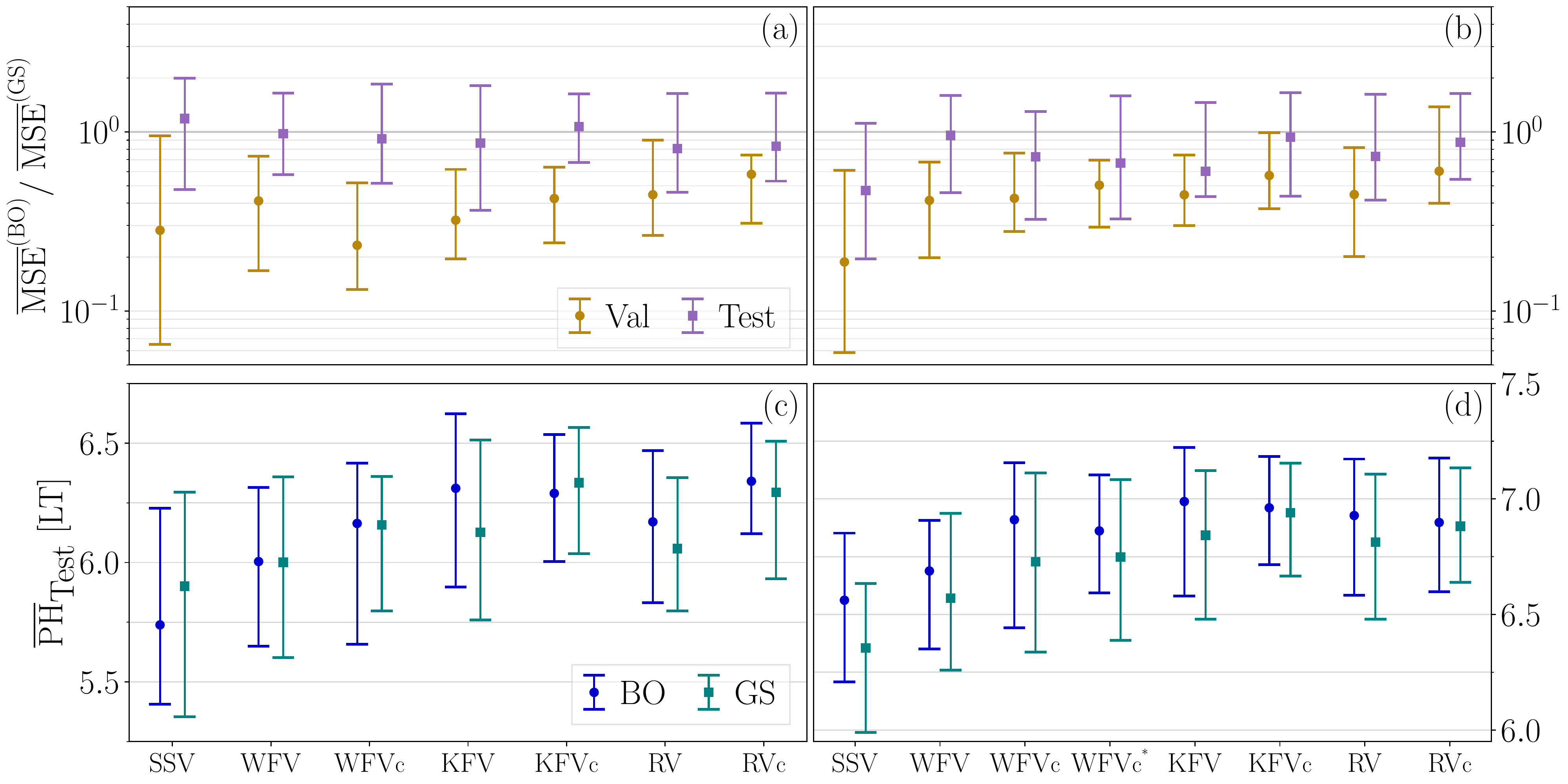}
    \caption{Comparison between hyperparameter optimization by Bayesian Optimization (BO) and Grid Search (GS).
    The performance metrics  are the Mean Square error (MSE) and Predictability Horizon (PH). 25th (lower bar), 50th (marker) and 75th (upper bar) percentiles. (a,c) short dataset, (b,d,) long dataset. 
    }
    \label{Lorenz_results}
\end{figure}

A comparison between Bayesian Optimization (BO) and Grid Search (GS) is shown in Fig. \ref{Lorenz_results}. 
Panels (a,b) show the ratio of the MSE between the optimal hyperparameters obtained by Bayesian Optimization and Grid Search in the validation and test sets. 
In both datasets, Bayesian Optimization outperforms  Grid Search in the validation set in $\sim75\%$ of the networks (except for one outlier). 
However, BO and GS perform similarly in the test set, especially in the  short dataset (a). 
In the long dataset (b),  Bayesian Optimization on average outperforms  Grid Search, although there is a  decrease in performance with respect to the validation set. 
Panels (c,d) show the Prediction Horizon (PH) in the test set. 
The chaotic K-Fold Validation and the chaotic Recycle Validation increase the Prediction Horizon by 0.5 LTs on average with respect to the Single Shot Validation. 
The Prediction Horizon of the long datasets (d) is  $\gtrsim0.5$ LTs larger than that of the short dataset (c). 
This results in the performance of the KFV$_{\mathrm{c}}$ and RV$_{\mathrm{c}}$ in the short dataset being closer to the performance of the SSV in the long dataset.
Because Bayesian Optimization does not produce a substantial increase in the Prediction Horizon with respect to  Grid Search, we conclude that the performance of the networks is more sensitive to the validation strategy rather than the optimization scheme.

\subsubsection{Model-informed ESN}

\begin{figure}[H]
    \centering
    \includegraphics[width=1.\textwidth]{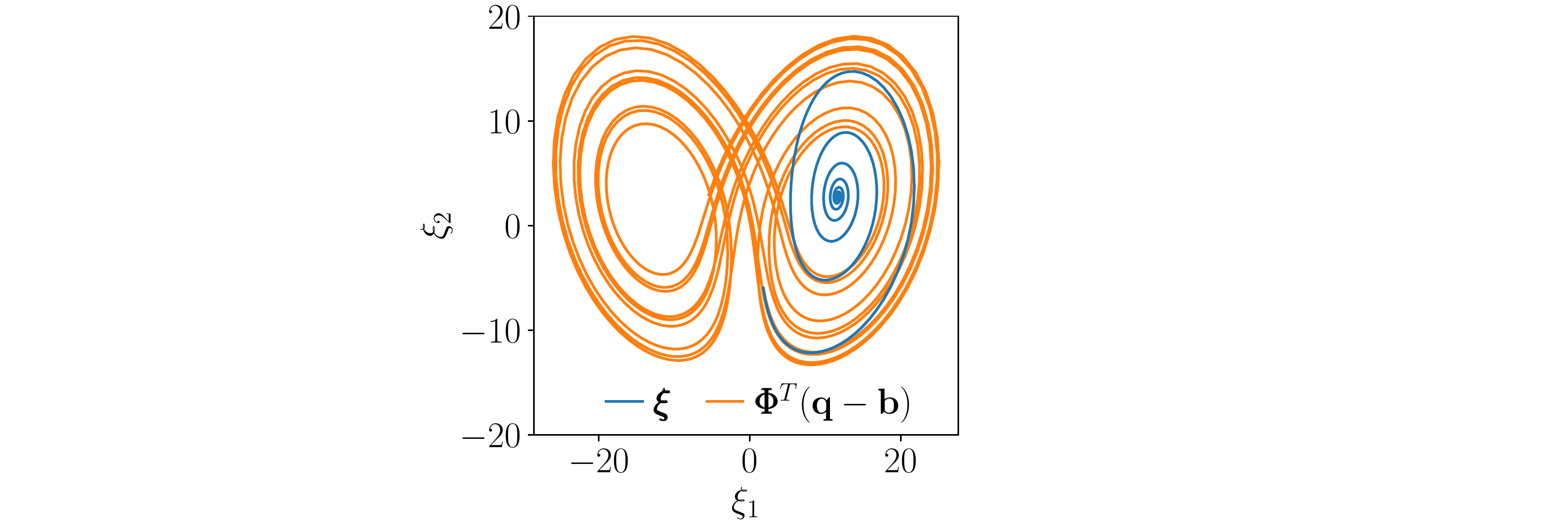}
    \caption{Projection of the trajectory onto the 2-dimensional model ($\pmb{\Phi}^T(\textbf{q-b})$) and autonomous evolution of the flat Galerkin approximation ($\pmb{\xi}$) in the POD-informed Echo State Network.
    }
    \label{Lorenz_POD}
\end{figure}

We leverage knowledge about the governing equations through $\pmb{\mathcal{K}}(\textbf{u}_{\mathrm{in}})$ in the model in Eq.~\eqref{eq:Hyb_ESN}. In this testcase, we use a reduced-order model obtained through Proper Orthogonal Decomposition (POD) \cite{lumley1967structure, weiss2019tutorial} to define a POD-informed ESN. POD provides a fixed rank subspace $\mathcal{E}$ of the state space, in which the projection of the original state vector optimally preserves its energy. The POD modes / energies are the eigenvectors / eigenvalues, of the data covariance matrix $\textbf{C}= \frac{1}{m-1}\textbf{U}^T\textbf{U}$.  
The $M\times N_{\mathrm{u}}$ matrix \textbf{U} is the vertical concatenation of the $M$ snapshots of the $N_{\mathrm{u}}$-dimensional timeseries used for washout, training and validation of the network, from which its mean, $\textbf{d} \in \mathbb{R}^{N_u}$, is subtracted columns-wise. We create an $N_{\mathrm{POD}}$-dimensional reduced-order model by taking the modes $\pmb{\phi}_i$ associated with the $N_{\mathrm{POD}}$ largest eigenvalues of $\textbf{C}$. Because $\textbf{C}$ is a symmetric matrix, its eigenvectors form an orthonormal basis, which is stored in the orthogonal matrix $\pmb{\Phi} = [\pmb{\phi}_1;...;\pmb{\phi}_{n_{\mathrm{POD}}}]$. The state vector $\textbf{q}$ is expressed as a function of its components $\pmb{\xi}$ in the subspace $\mathcal{E}$ spanned by $\pmb{\Phi}$, and its components $\pmb{\eta}$ in the orthogonal complement of $\mathcal{E}$ spanned by the basis $\pmb{\Psi}$ \begin{align}\textbf{q} = \pmb{\Phi}\pmb{\xi}+\pmb{\Psi}\pmb{\eta}+\textbf{d}.\end{align}
The evolution equations are then obtained by using a \emph{flat Galerkin approximation} \cite{matthies2003nonlinear}, which neglects the contribution of the orthogonal complement: $\pmb{\Psi}\pmb{\eta}\simeq0$. The nonlinear dynamical system $\bf{\dot{q}}=\textbf{f(q)}$ is projected onto $\mathcal{E}$ through $\pmb{\xi} = \pmb{\Phi}^T(\mathbf{q-d})$ as 
\begin{gather}
    \dot{\pmb{\xi}} = \pmb{\Phi}^T\textbf{f}(\pmb{\Phi}\pmb{\xi}+\textbf{d})
    \label{eq:Galerkin}
\end{gather}
In the POD-informed ESN model ($\textbf{q}\equiv\textbf{u}_{\mathrm{in}}$), we use $N_{\mathrm{POD}}=2$ to generate the reduced-order model, which accounts for 96$\%$ of the energy of the original signal. We use the evolution of the trajectory on the POD subspace, $\mathcal{E}$, to inform the ESN through $\pmb{\mathcal{K}}(\textbf{u}_{\mathrm{in}}(t_i)) = \pmb{\xi}(t_{i+1})$. We solve the ODE system in Eq.~\eqref{eq:Galerkin} using at each time step forward Euler with initial condition $\pmb{\xi}(t_{i}) = \pmb{\Phi}^T(\textbf{u}_{\mathrm{in}}(t_i)-\textbf{d})$, which is the projection of the input to the network onto $\mathcal{E}$. The projection of the trajectory and the autonomous evolution of the flat Galerkin approximation are shown in Fig. \ref{Lorenz_POD}.
The reduced order model dynamics, $\pmb{\xi}$, differ significantly from the dynamics of the entire state, $\pmb{\Phi}^T(\textbf{q-b})$. However, we show in the next paragraph that embedding model knowledge, yet imperfect, can improve the performance of the networks. 

\begin{figure}[H]
    \centering
    \includegraphics[width=1.\textwidth]{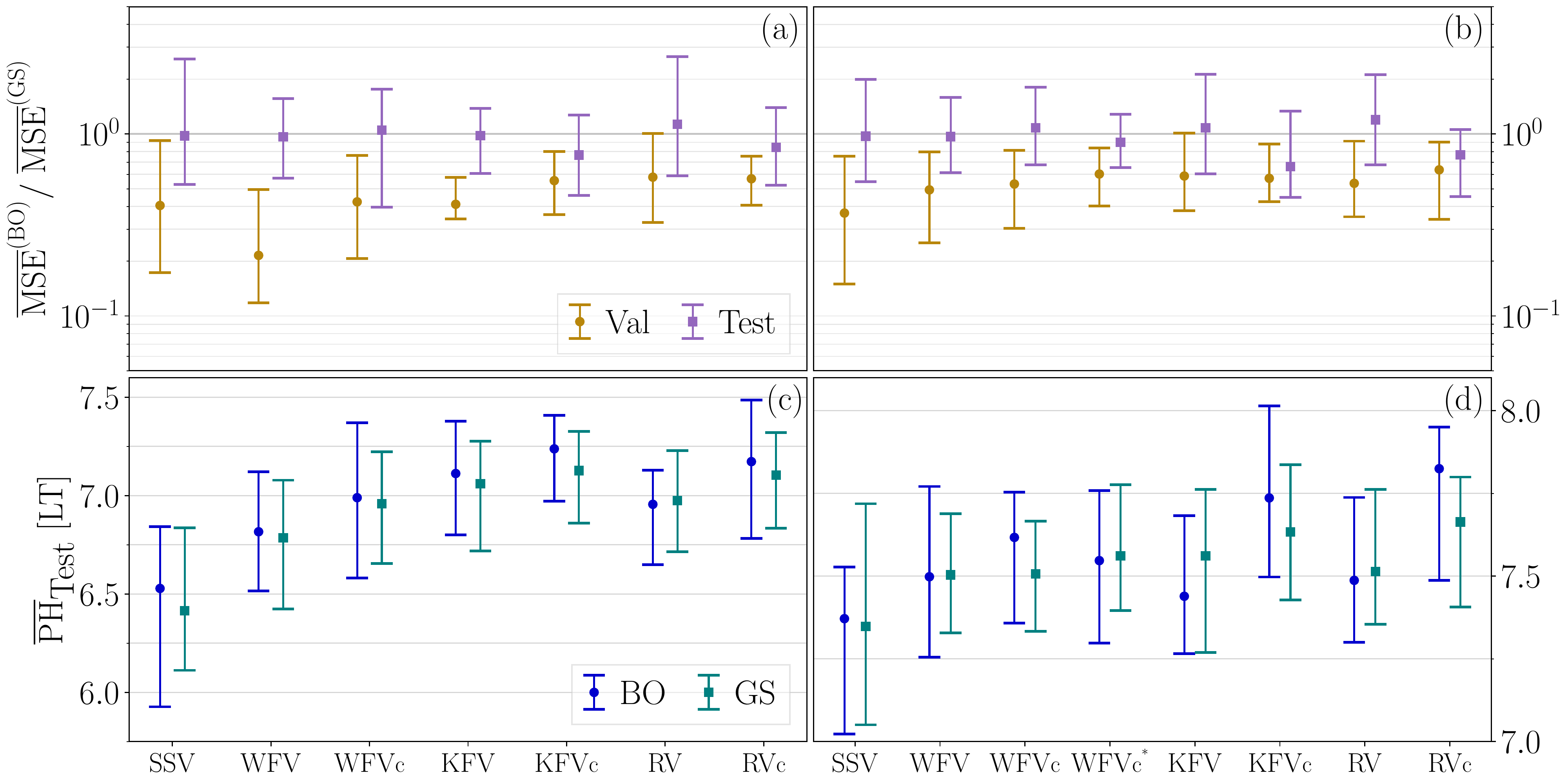}
        \caption{Same as Fig.~\ref{Lorenz_results} for the POD-informed ESN.
    }
    \label{Hyb_Lorenz_results}
\end{figure}

As compared to the model-free ESN, there is a decrease in correlation between the validation and test sets for almost all the validation strategies (Tab. \ref{tab:Hyb_24LT_12LT}). 
The Single Shot Validation is still outperformed by the other strategies, while the chaotic K-Fold Validation and chaotic Recycle Validation have the highest correlation. 
Panels (a,b) of Fig. \ref{Hyb_Lorenz_results} show similar results to the model-free case:  Bayesian Optimization outperforms  Grid Search in the validation set, but the two schemes perform similarly in the test set. 
The only exception are the chaotic K-Fold Validation and chaotic Recycle Validation in the long dataset (b), in which  Bayesian Optimization outperforms  Grid Search for up to 75\% of the networks in the test set. 
Panels (c,d) show that embedding knowledge of the governing equation produces an increase of 1LT in the Prediction Horizon with respect to the model-free case (see Fig. \ref{Lorenz_results}). 
The qualitative behaviour of the validation strategies remains similar to the model-free case. 
To conclude, although the POD-informed architecture  does increase the performance, it does not increase the robustness of the networks with respect to the model-free ESN. 
\begin{table}[H]
\centering
    \caption{Spearman  coefficients between validation and test sets. Bold text indicates the highest correlation in the dataset. }
\begin{tabular}{c c c c c c c c c c}
$\Tilde{r}_S$ & $\,$ &  SSV & WFV &  WFV$_{\mathrm{c}}$ &  WFV$_{\mathrm{c}}^*$ & KFV & KFV$_{\mathrm{c}}$ & RV & RV$_{\mathrm{c}}$ \\
\toprule
Short dataset  (12LTs) & & 0.15 & 0.39 & 0.34  & - & 0.32 & \textbf{0.73} & 0.41 & 0.56\\
\midrule 
Long dataset  (24LTs) & & 0.19 & 0.42 & 0.36 & 0.51 & 0.59 & \textbf{0.80} & 0.55 & \textbf{0.80} \\
\toprule
\end{tabular}
\label{tab:Hyb_24LT_12LT}
\end{table}

\section{Validation for quasiperiodic solutions}

\label{sec:Oscillator}

\begin{figure}[H]
    \centering
    \includegraphics[width=1.0\textwidth]{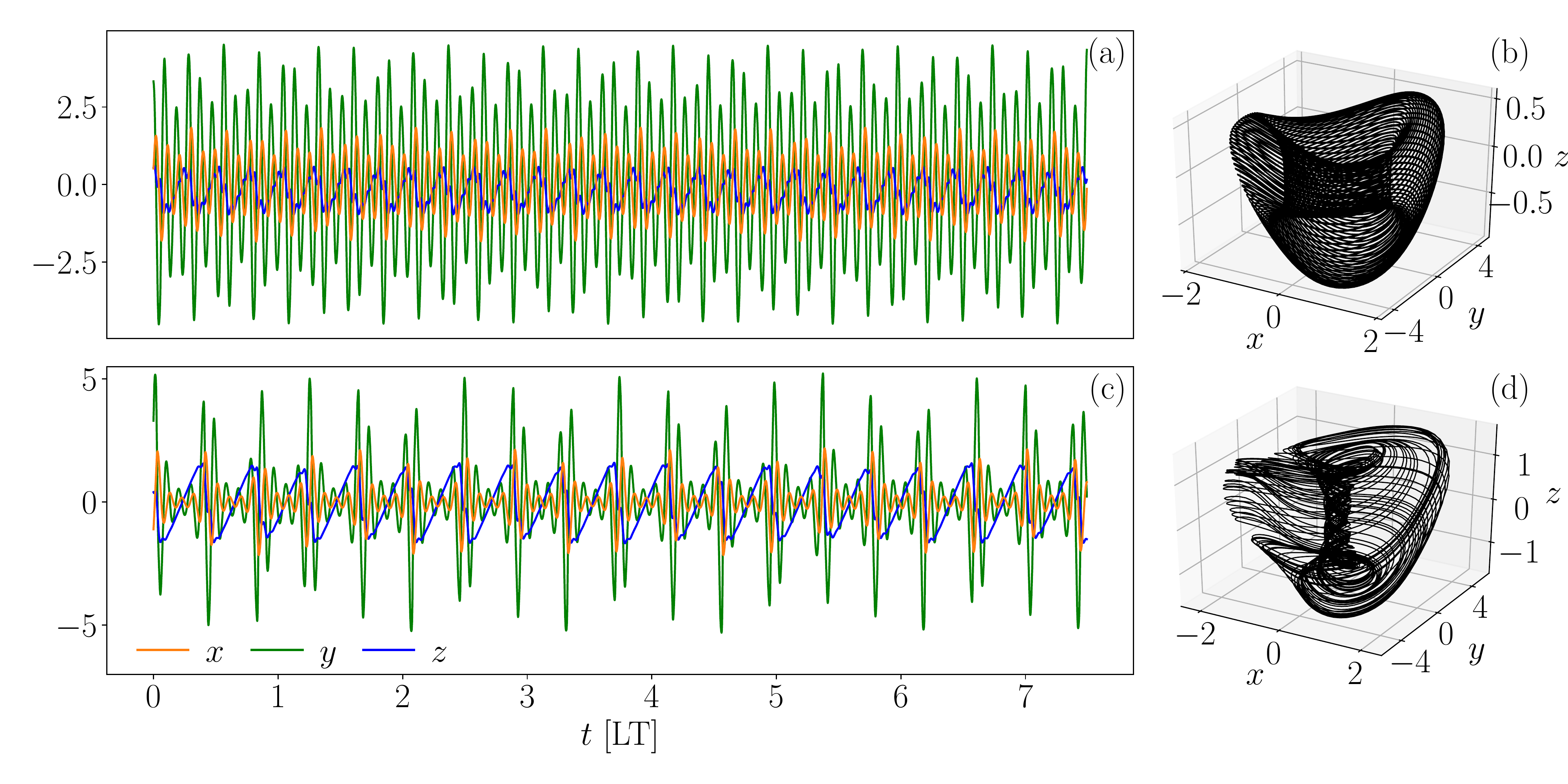}
    \caption{Kuznetsov oscillator. (a) Quasiperiodic and (c) chaotic time series; (b) quasiperiodic and (d) chaotic phase plots for a longer time window. Time is expressed in the Lyapunov time (LT)  of the chaotic case ($\textrm{LT}\approx 25$ \cite{kuznetsov2010simple}).
    }
    \label{QP_Ch_TS}
\end{figure}

We analyse the nonlinear oscillator proposed by Kuznetsov et al. \cite{kuznetsov2010simple}, which physically represents a self-oscillatory discharge in an electric circuit. 
The oscillator is a three-dimensional system, which can display periodic, quasiperiodic and chaotic behaviours as a function of the parameters $[\lambda, \omega_0, \mu]$
\begin{align}
    \dot{x} &= y, \nonumber \\
    \dot{y} &= y(\lambda + z + x^2 - \frac{1}{2}x^4) - \omega_0^2x, \nonumber \\
    \dot{z} &= \mu - x^2.
    \label{Oscillator}
\end{align}
The primary purpose of this testcase is to compare the  robustness of Echo State Networks in forecasting quasiperiodic solutions versus chaotic solutions. 
This enables us to determine whether the challenges encountered in the Lorenz system are specific to learning chaotic time series. 
We obtain quasiperiodic and chaotic solutions by setting $\lambda=0,\; \omega_0=2.7$ and $\mu_{Qp}=0.9$ and $\mu_{Ch}=0.5$, as shown in Figs.~\ref{QP_Ch_TS}(b,d), respectively. 
(For completeness, in this section, we report the Kuznetsov chaotic solution as well.) 
The datasets of 7.5 LTs that we use for washout, training and validation are shown in panels (a,c).

\subsection{Hyperparameter optimization}
The network parameters, the size of the ensemble, and the optimization strategies are the same as those of section \ref{sec:SSV_lor}. 
We modify the input scaling, $b_{\mathrm{in}}=0.1$, for it to have the same order of magnitude of the input, which is obtained by normalizing the signal by its maximum variation component-wise. 
We study the enlarged interval $\rho=[0.01,1]$ because we observed empirically that the optimal hyperparameters often lie in the range $\rho \leq 0.1$. 
Given the multiple orders of magnitude of the spectral radius, the hyperparameter space is analysed in a logarithmic scale. 
We use the same architecture and validation strategies for the quasiperiodic and chaotic case (as detailed in the Supplementary Material, S.1). 
The different strategies are tested by computing the arithmetic mean $\overline{\textrm{PH}}_{\textrm{test}}$ of the Prediction Horizon on $N_t$ starting points for the chaotic case, and by computing the geometric mean $\overline{\textrm{MSE}}_{\textrm{test}}$ of the Mean Squared Error in 2 LTs intervals starting from the same points. 
In the chaotic case, we select $N_t=75$, whereas in the quasiperiodic we select $N_t=50$ through the procedure described in \ref{A:Convergence}. 
The performance in the quasiperiodic dataset is assessed only through the Mean Squared Error  because the Prediction Horizon is  infinite, i.e., a quasiperiodic solution has zero dominant Lyapunov exponents~\cite{kantz2004nonlinear}.

\subsubsection{Model-free ESN}
\begin{figure}[H]
    \centering
    \includegraphics[width=1.\textwidth]{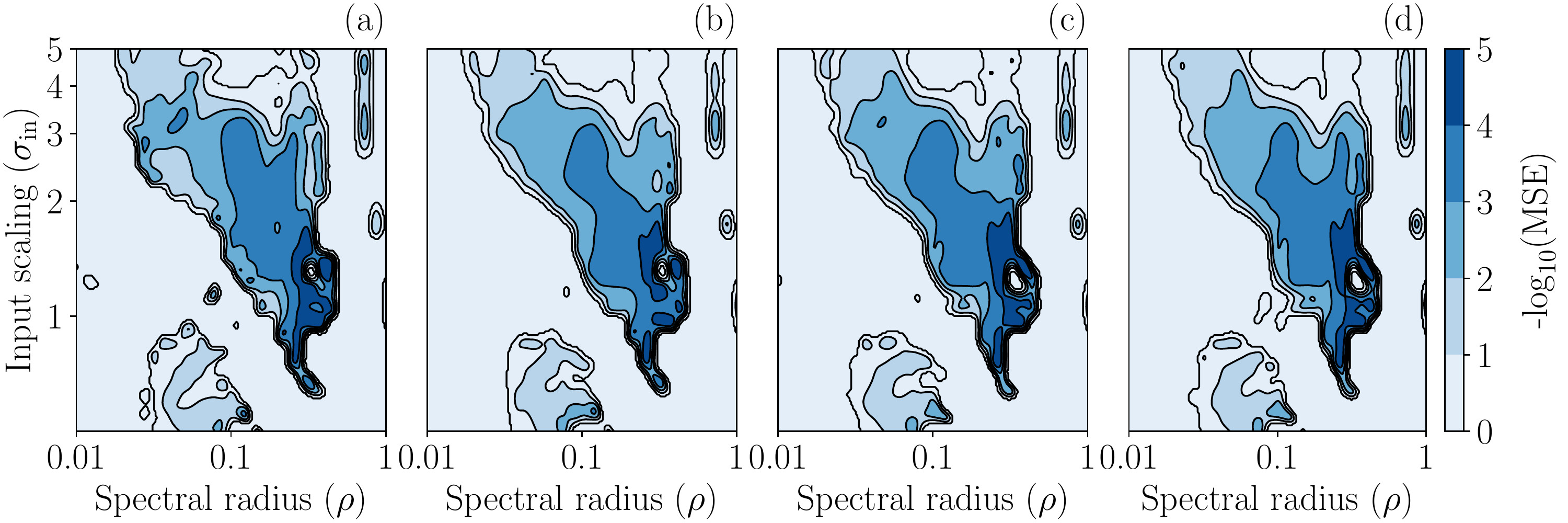}
    \caption{Mean of the Gaussian Process reconstruction for the quasiperiodic dataset for a representative network of the ensemble. Validation set for (a) Single Shot Validation, (b) chaotic K-Fold Validation, and (c) chaotic Recycle Validation; and (d) test set. The $\mathrm{MSE}$ is saturated to be $\leq 1$.  The Gaussian Process is based on a grid of 30$\times$30 data points.
    }
    \label{Osc_Strat_MSE_Surfaces}
\end{figure}

\begin{figure}[H]
    \centering
    \includegraphics[width=1.\textwidth]{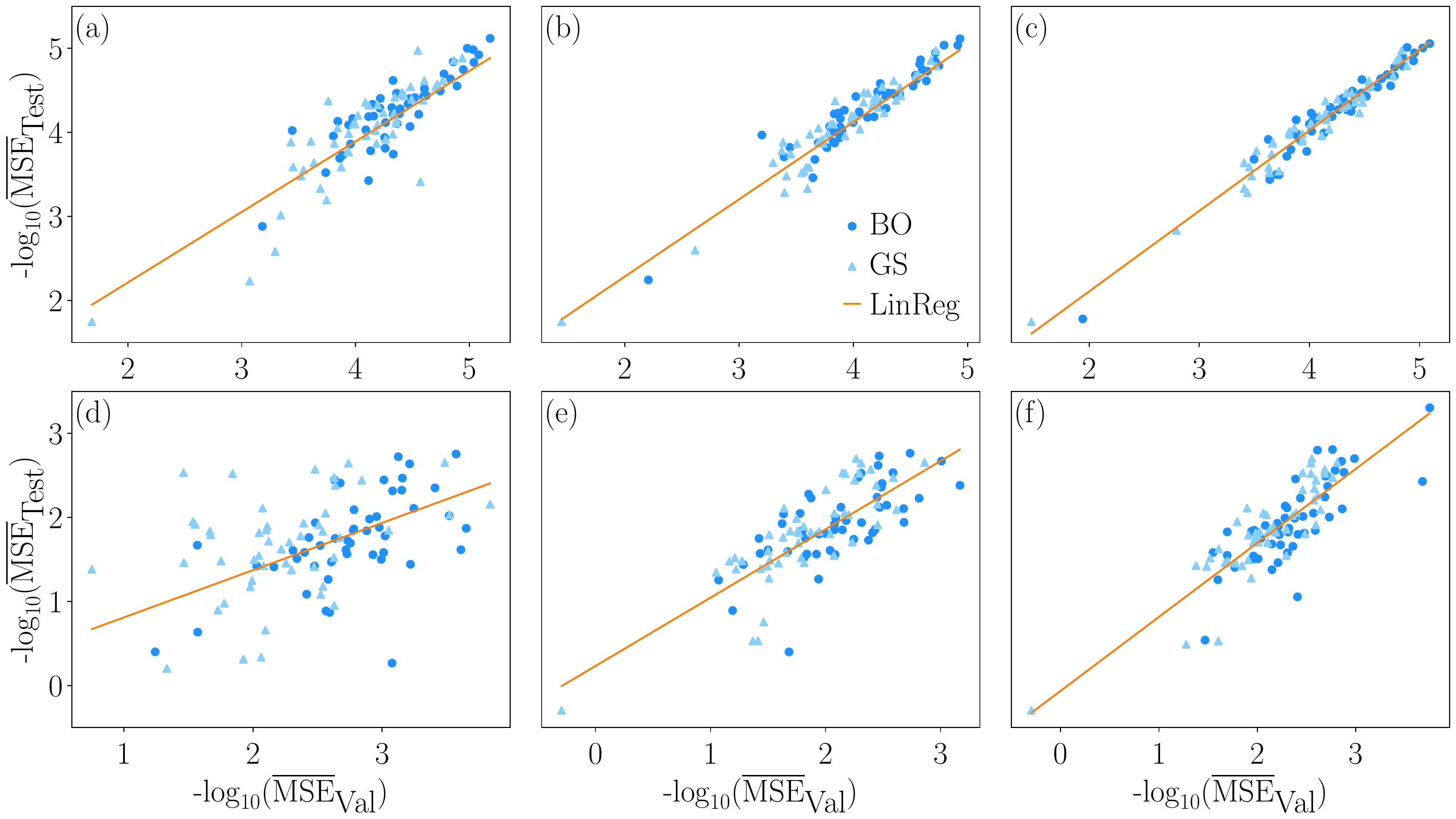}
    \caption{Linear regression (LinReg) and scatter plot of the MSE of the optimal hyperparameters obtained from Bayesian Optimization (BO) and Grid Search (GS) for each network. 
    (a,d) Single Shot Validation, (b,e) chaotic K-Fold Validation and (c,f) chaotic Recycle Validation. (a-c) quasiperiodic and (d-f) chaotic datasets.
    }
    \label{Oscillator_TestVal_scatter}
\end{figure}

Figure \ref{Osc_Strat_MSE_Surfaces} shows the MSE in the hyperparameter space for the quasiperiodic case. 
The plots for three validation strategies, (a-c), are very close to the MSE in the test set, (d), which means that hyperparameters that perform well in the validation set, perform as well in the test set. 
This is in contrast with the behaviour in chaotic solutions (see Fig. \ref{Strat_MSE_Surfaces}). 
The Spearman coefficients (Tab. \ref{tab:QP_Ch}) confirm that the correlation between validation and test sets is  higher in the quasiperiodic dataset than the chaotic dataset. 
Notably, the peak $\Tilde{r}_s=0.97$ obtained in the Recycle Validations indicates almost complete correlation. 
As before, the Single Shot Validation is outperformed by the K-Fold Validation and Recycle Validation, but its correlation in the quasiperiodic dataset is higher than that of chaotic cases. 
The high correlation in the quasiperiodic dataset is identified as the dense clustering around the linear regression of Fig. \ref{Oscillator_TestVal_scatter}. 
Two remarks can be made. 
On the one hand, the high correlation in the quasiperiodic dataset implies that the challenges in producing robust results in Echo State Networks in chaotic attractors are due to the complexity of the chaotic signal, rather than the properties of the networks.
On the other hand, the marked difference in performance between different networks is still present in the quasiperiodic dataset, which means that ESNs are sensitive to the realizations (further analysis is reported in \ref{sec: A_fix}). 
Practically, we advise that different networks  be optimized independently in the quasiperiodic case as well.

\begin{table}[H]
\centering
    \caption{Spearman  coefficients between validation and test sets. Bold text indicates the highest correlation in the dataset. }
\begin{tabular}{c c c c c c c c c}
$\Tilde{r}_S$ & $\,$ &  SSV & WFV &  WFV$_{\textrm{c}}$ & KFV & KFV$_{\textrm{c}}$ & RV & RV$_{\textrm{c}}$ \\
\toprule
Quasiperiodic dataset & & 0.80 & 0.75 & 0.71 & 0.93 & 0.92 & \textbf{0.97} & \textbf{0.97}\\
\midrule 
Chaotic dataset & & 0.49 & 0.48 & 0.58 & 0.70 & 0.76 & 0.66 & \textbf{0.81} \\
\toprule
\end{tabular}
\label{tab:QP_Ch}
\end{table}
 
\begin{figure}[H]
    \centering
    \includegraphics[width=1.0\textwidth]{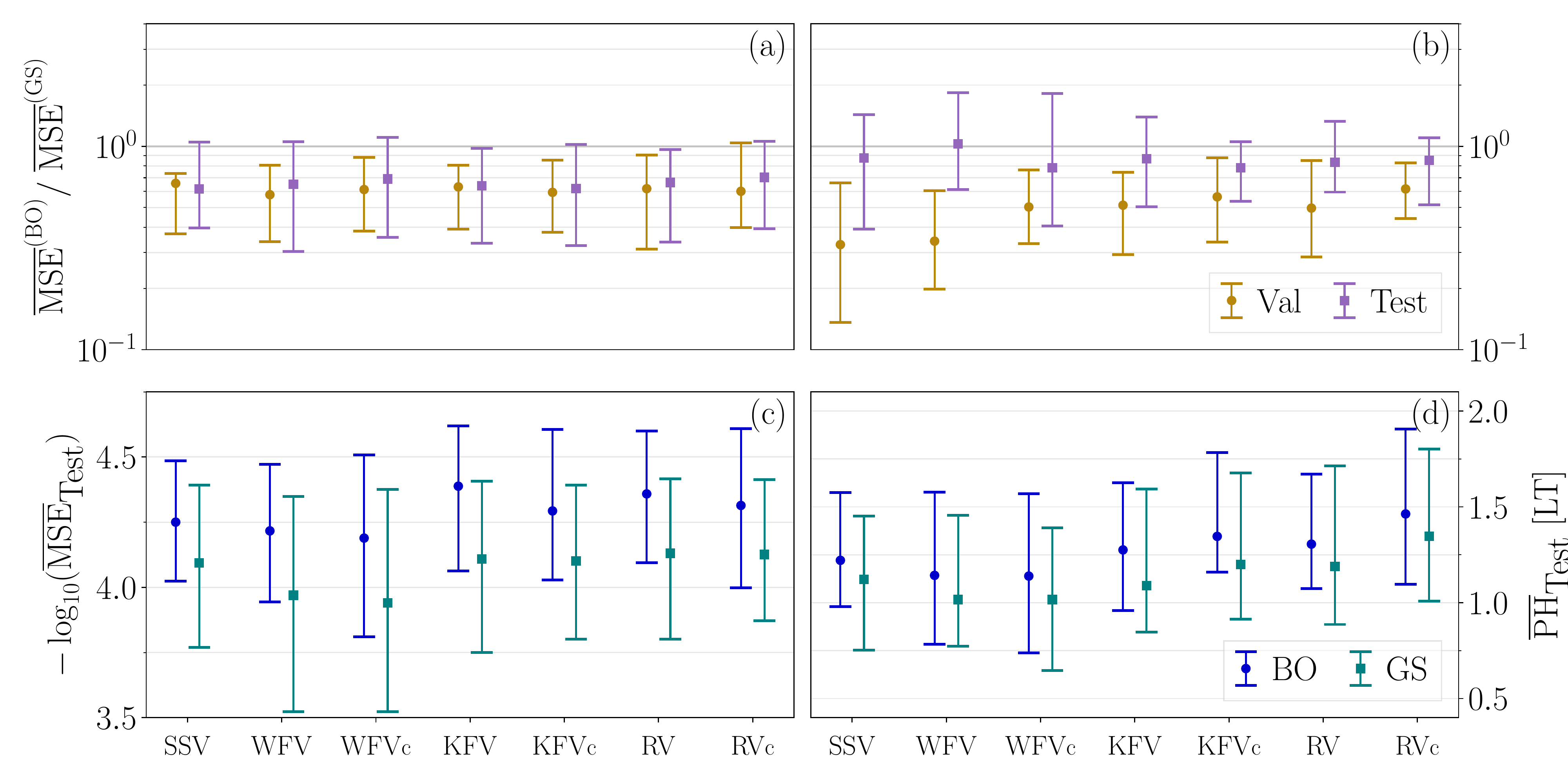}
    \caption{Comparison between hyperparameter optimization by Bayesian Optimization (BO) and Grid Search (GS) for the two performance metrics (MSE, PH). 25th (lower bar), 50th (marker) and 75th (upper bar) percentiles. (a,c) quasiperiodic dataset, (b,d,) chaotic dataset. 
    }
    \label{Oscillator_Results}
\end{figure}

Panels (a,b) of Fig. \ref{Oscillator_Results} show the ratio of the MSE between the optimal hyperparameters obtained by Bayesian Optimization (BO) and the optimal hyperparameters from Grid Search (GS) in the validation and test sets. 
On the one hand, in the quasiperiodic case (a) the performance in the validation set is similar to the test set.
One the other hand, in the chaotic case (b)  BO outperforms GS in the validation set, although the two schemes perform similarly in the test set. 
In panels (c,d), we show the performance of the networks in the test set using the MSE for the quasiperiodic dataset (c) and the Prediction Horizon in the chaotic dataset (d). 
In the quasiperiodic dataset, Bayesian Optimization  outperforms  Grid Search, and the K-Fold Validations and Recycle Validations outperform the other validation strategies. 
In the chaotic dataset, as seen in the Lorenz system, Bayesian Optimization only slightly outperforms Grid Search, while the K-fold Validations and Recycle Validations still outperform the other validation strategies.

\subsubsection{Model-informed ESN}
We design a Forward Euler (FE) informed ESN \eqref{eq:Hyb_ESN} by integrating in time with forward Euler the $y$ equation \eqref{Oscillator} only 
\begin{equation}
    \pmb{\mathcal{K}}(\mathbf{u}_{\mathrm{in}}) = y +  dt(y(\lambda + z + x^2 - \frac{1}{2}x^4) - \omega_0^2x).
\end{equation}

\begin{figure}[H]
    \centering
    \includegraphics[width=1.0\textwidth]{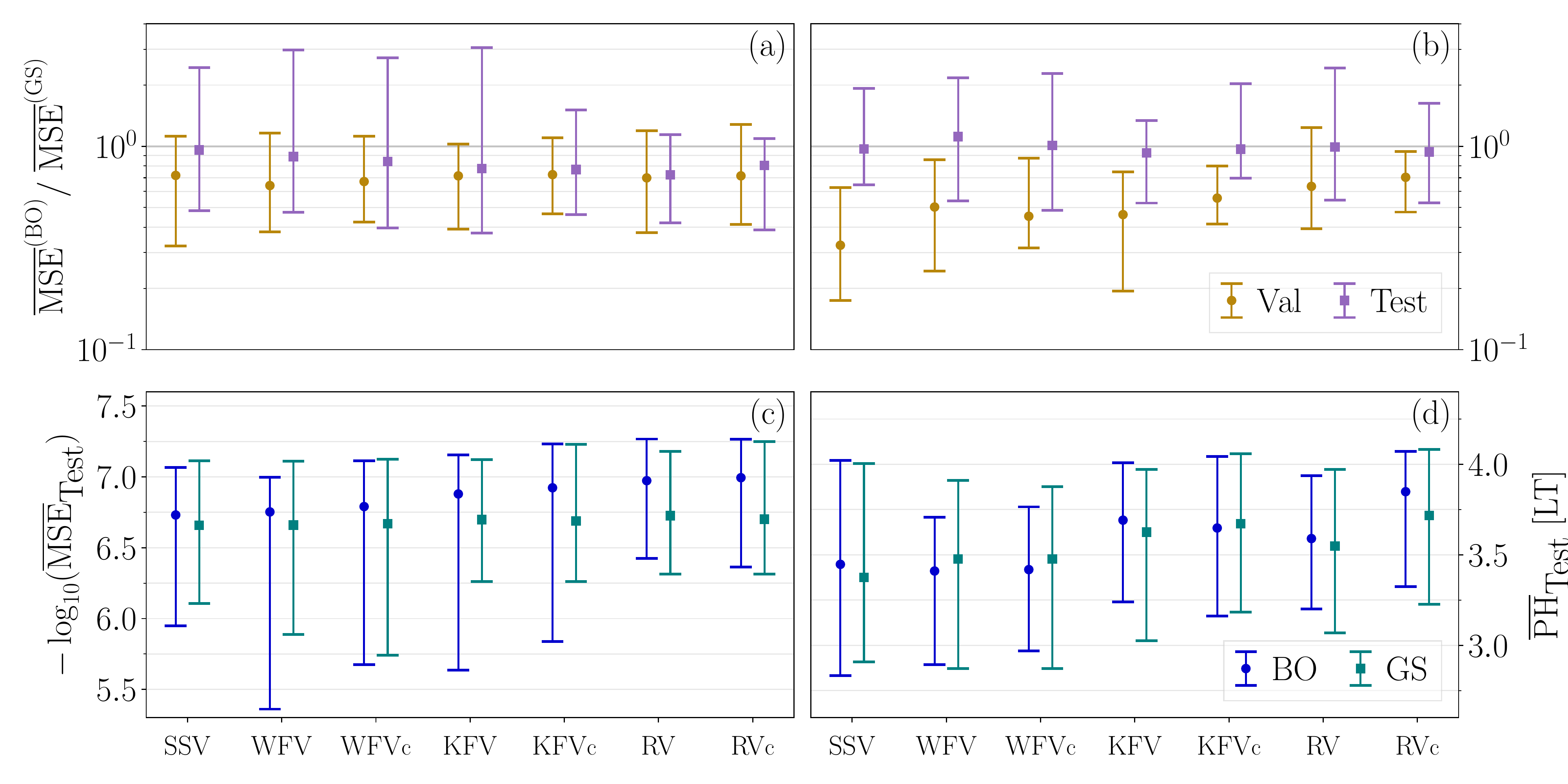}
    \caption{Same as Fig.~\ref{Oscillator_Results} for the FE-informed ESN.}
    \label{Hyb_Oscillator_Results}
\end{figure}

Tab. \ref{tab:Hyb_QP_Ch} shows the Spearman coefficients for the FE-informed model. 
In the quasiperiodic dataset, the correlation decreases for all the validation strategies with respect to the model-free case (see Tab. \ref{tab:QP_Ch}) except for the Recycle Validations, which have the highest correlation. However, in the chaotic dataset, the correlation increases for all the validation strategies. 
Here, the chaotic K-Fold Validation and chaotic Recycle Validation are the strategies with the highest correlation.
Fig.\ref{Hyb_Oscillator_Results}(a) shows that the decrease in correlation in the quasiperiodic dataset causes Bayesian Optimization to generate larger MSE than Grid Search with respect to the model-free case. 
In panel (b), we observe that there is still a marked discrepancy between the performance of the optimization schemes in the validation and test sets. 
Panels (c,d) show the performance of the FE-informed ESN in the test set.
In both datasets, the performance improves when leveraging knowledge about the governing equations: the MSE decreases by about two orders of magnitude, and the Prediction Horizon improves by $\gtrsim 2$ Lyapunov Times with respect to the model-free case (see Fig. \ref{Oscillator_Results}). 
The improvement in performance, however, does not correspond to a consistent increase in correlation between validation and test sets. 
The performance of Bayesian Optimization with respect to Grid Search in the test set does not necessarily improve. 
In the same fashion as the Lorenz system, the FE-informed architecture {\it per se} does enhance the performance, but it does not enhance the robustness of Echo State Networks.

\begin{table}[H]
\centering
    \caption{Spearman  coefficients between validation and test sets. Bold text indicates the highest correlation in the dataset. }
\begin{tabular}{c c c c c c c c c}
$\Tilde{r}_S$ & $\,$ &  SSV & WFV &  WFV$_{\textrm{c}}$ & KFV & KFV$_{\textrm{c}}$ & RV & RV$_{\textrm{c}}$ \\
\toprule
Quasiperiodic dataset & & 0.78 & 0.65 & 0.67  & 0.71 & 0.80 & \textbf{0.98} & \textbf{0.98}\\
\midrule 
Chaotic dataset & & 0.57 & 0.63 & 0.63 & 0.75 & 0.79 & 0.71 & \textbf{0.85} \\
\toprule
\end{tabular}
\label{tab:Hyb_QP_Ch}
\end{table}

\section{Conclusions}
\label{sec:Conclusion}
The Echo State Network (ESN) is a reservoir computing architecture that is able to learn accurately the nonlinear dynamics of systems from data. 
The overarching objective of this paper is to investigate and improve the robustness of ESNs, with a focus on the forecasting of chaotic systems. 
First, we analyse the Single Shot Validation, which is the commonly used strategy to select the hyperparameters.
We show that the Single Shot Validation is the least performing strategy to fine-tune the hyperparameters. 
Second, we validate the ESNs on multiple points of the chaotic attractor, for which the validation set is not necessarily subsequent in time to the training set. 
We propose the Recycle Validation and the chaotic version of existing validation strategies based on multiple folds, such as the Walk Forward Validation and the K-Fold Cross Validation. 
The K-Fold Validation and Recycle Validation  offer the greatest robustness and performance, with their chaotic versions outperforming the corresponding  regular versions. 
Importantly, the Recycle Validation is computationally cheaper than the K-Fold Cross Validation. 
Third, we compare Bayesian Optimization with Grid Search to compute the optimal hyperparameters. 
We find that Bayesian Optimization is an optimization scheme that consistently finds a set of hyperparameters that perform significantly better than the Grid Search in the validation set. 
On the one hand, in learning quasiperiodic solutions, hyperparameters that work optimally in the validation set continue to work optimally in the test set. 
This is because quasiperiodic solutions are predictable (i.e., they do not have positive Lyapunov exponents). 
This finding is, thus, expected to generalize to other predictable solutions, such as frequency-locked solutions and limit cycles. 
On the other hand, in learning chaotic solutions, hyperparameters that work optimally in the validation set do not necessarily work optimally in the test set. 
We argue that this occurs because of the chaotic nature of the attractor, in which the nonlinear dynamics, although deterministic, manifest themselves as unpredictable variations. 
Fourth, we analyse the model-free ESN, which is fully data-driven, and the model-informed ESN, which leverages knowledge of the governing equations. 
We find that the model-informed architecture markedly improves the network's prediction capabilities, but it does not improve the robustness. 
Finally, we find that the optimal hyperparameters are significantly sensitive to the random initialization of the ESN. 
Practically, when working with an ensemble of ESNs, we recommend computing the optimal hyperparameters for each network. 
In the test performed in the paper, this can increase up to six Lyapunov Times the network's Prediction Horizon as compared to using the same set of hyperparameters for all realizations.\\

This work opens up new possibilities for using Echo State Networks and, in general, recurrent neural networks, for robust learning  of chaotic dynamics.  

\section*{Acknowledgements}
A. Racca is supported by the EPSRC-DTP and the Cambridge Commonwealth, European \& International Trust under a Cambridge European Scholarship. L. Magri is supported by the Royal Academy of Engineering Research Fellowship scheme and the visiting fellowship at the Technical University of Munich – Institute for Advanced Study, funded by the German Excellence Initiative and the European Union Seventh Framework Programme under grant agreement
n. 291763. The authors would like to thank Dr. N. A. K. Doan and F. Huhn for  insightful discussions.

\appendix
\section{Correlation between the mean-squared error and predictability horizon}
\label{A:Val_Fig}

\begin{figure}[H]
    \centering
    \includegraphics[width=1.\textwidth]{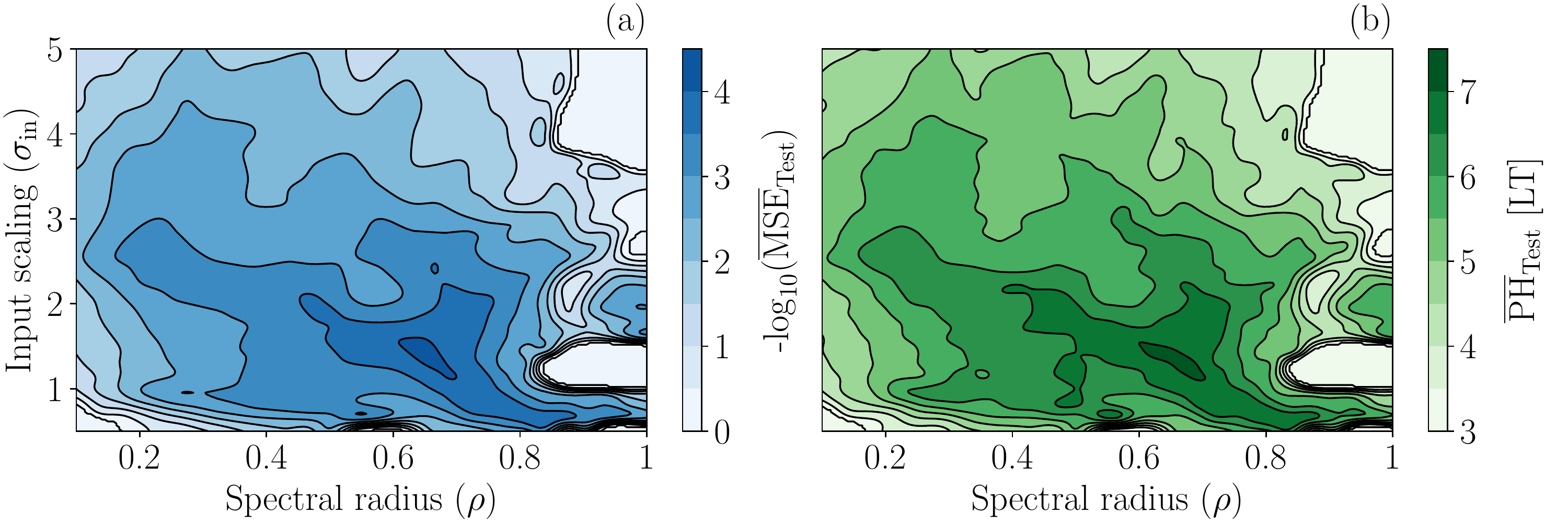}
    \caption{Mean of the Gaussian Process reconstruction from a 900 points grid in the test set for (a) $\log_{10}(\mathrm{MSE})$ and (b) Prediction Horizon (PH) for a representative network in the short dataset. For visualization purposes we saturate the $\mathrm{MSE}$ to be $\leq 1$,  and the PH to be $\geq 3$.
    The MSE and PH closely resemble one another.
    }
    \label{PH_MSE_surf}
\end{figure}

\begin{figure}[H]
    \centering
    \includegraphics[width=1.0\textwidth]{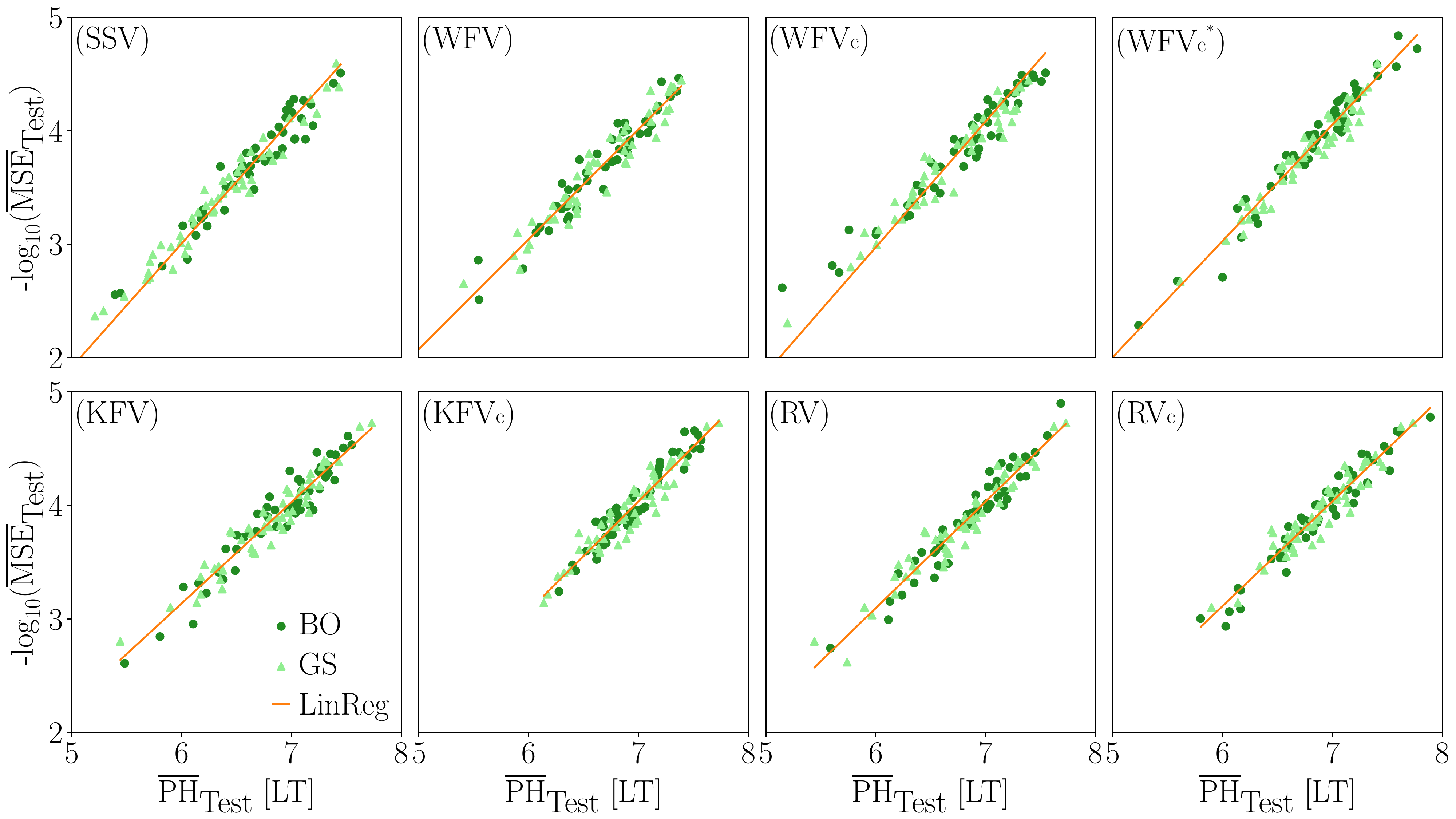}
    \caption{Linear regression (LinReg) and scatter plot for the test set MSE and Prediction Horizon in the long dataset of the optimal hyperparameters from Bayesian Optimization (BO) and Grid Search (GS) for the Single Shot Validation (SSV), Walk Forward Validation (WFV), K-Fold Validation (KFV), Recycle Validation (RV), and their chaotic versions, with subscript $c$.  The trends are highly correlated. 
    }
    \label{MSE_PH_Correlations}
\end{figure}

We show the high correlation between the Mean Squared Error and the Predictability Horizon given the same starting point for prediction. 
Figure~\ref{PH_MSE_surf} shows the Gaussian Process reconstruction from 900 (30$\times$30) grid points in the hyperparameter space in the $N_t=100$ test set (\ref{A:Convergence}) of the Lorenz system for a representative network realization. 
The two quantities show almost identical behaviour. 
Figure~\ref{MSE_PH_Correlations} shows the scatter plots for the Prediction Horizon and the Mean Square Error in the test set for the optimal hyperparameters for the ensemble in the Lorenz system. The two quantities are highly correlated, with a Spearman coefficient, $r_s \geq 0.95$, for all the validation strategies.  

\section{Computational time}

\label{A_CompTime}

\begin{figure}[H]
    \centering
    \includegraphics[width=1.\textwidth]{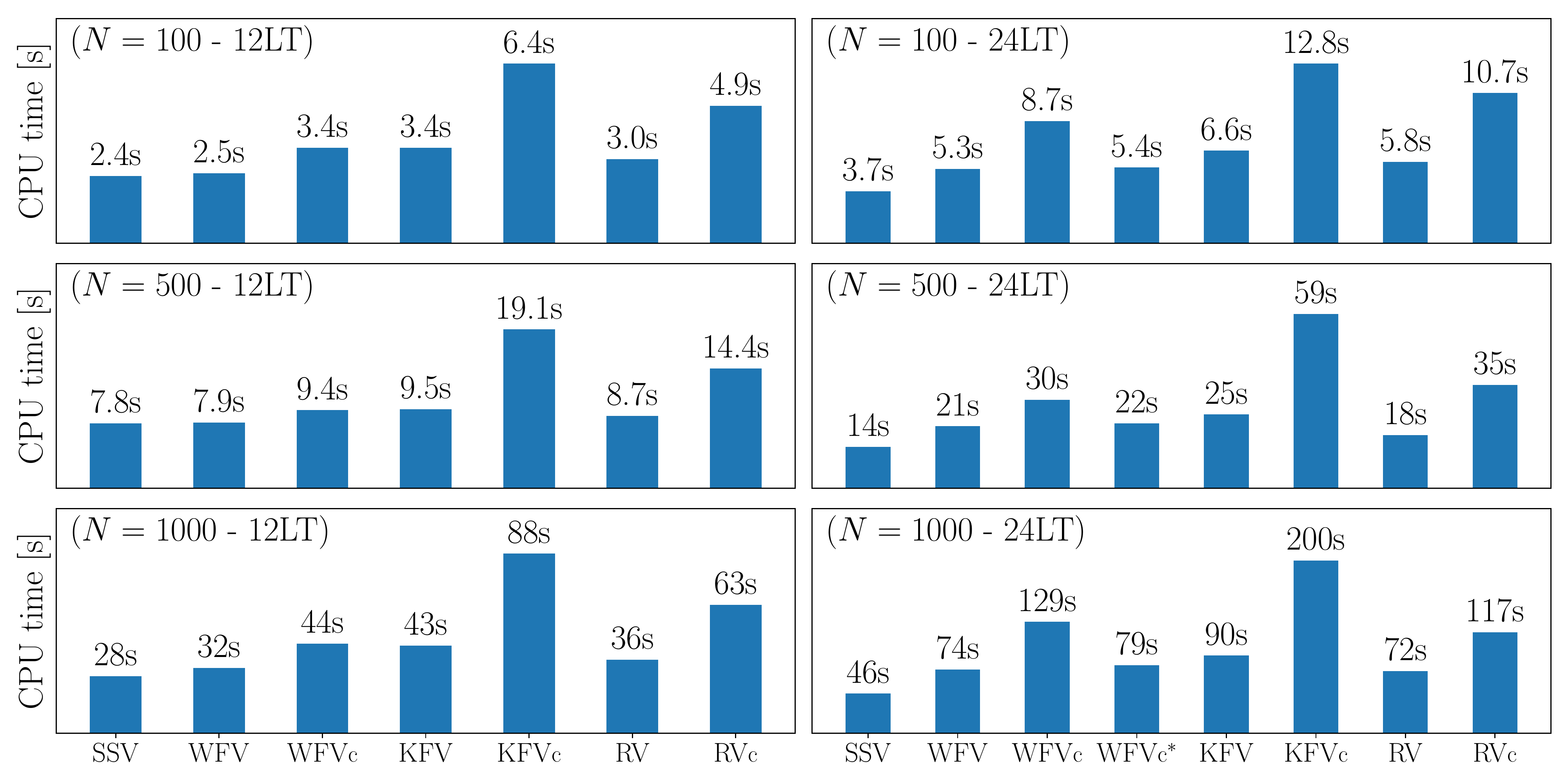}
    \caption{CPU time required for a single network of size $N$ to perform a 7$\times$7 Grid Search in hyperparameters space  in the 12LT and 24LT datasets of the Lorenz system. The validation strategies are the Single Shot Validation (SSV), the Walk Forward Validation (WFV), K-Fold Validation (KFV), Recycle Validation (RV), and respective chaotic versions (subscript $c$). The runs are on a single Intel i7-8750H processor.}
    \label{Lorenz_CompTime}
\end{figure}

In Fig. \ref{Lorenz_CompTime}, we show the CPU time required by the validation strategies to perform a Grid Search in hyperparameters space for a single network. The computational advantage of the Recycle Validation increases with the size of the dataset and the size of the reservoir. We expect the improvement in computational time to be more significant in RNN architectures whose training is more expensive, such as LSTMs and GRUs.

The Bayesian Optimization described in section \ref{sec:SSV_lor} costs approximately 6 seconds more per network in all the cases shown. This because the additional cost of the Bayesian Optimization is independent of the cost of the evaluation function. 

\section{Bayesian Optimization for hyperparameters}
\label{A_BO}
After we evaluate the objective function at $N_{st}$ starting points, the  objective function is reconstructed in the hyperparameter search space using the function evaluations as data points for noise-free Gaussian Process Regression. 
 The computational cost of the regression is proportional to $N_d^3$, where $N_d$ is the number of data points, because of the inversion of the covariance matrix. 
The inversion is performed by  Cholesky factorization regularized by the addition of $\alpha=10^{-10}$ on the diagonal elements. 

Once the Gaussian Process is performed, the next point at which to evaluate the objective function is selected in the hyperparameter space to maximize the acquisition function. 
The acquisition function evaluates a potential point usefulness in finding the global minimum, so that points with a high value of the acquisition function are selected during the search. 
A new point can be chosen for one of two reasons: 
(i) to try to find a new minimum by using current knowledge of the search space and 
(ii) to increase the knowledge of the space by exploring new regions. 
This trade-off is called balance between exploitation and exploration. 
Practically, the most used acquisition functions in the literature are the Probability of Improvement (PI), the Expected Improvement (EI) and the Lower Confidence Bound (LCB) \cite{brochu2010tutorial}. 
On a given testcase, it is difficult to determine a priori which acquisition function will perform better. 
For this reason we use the gp-hedge algorithm~\cite{hoffman2011portfolio}, which improves the performance with respect to the single acquisition functions.
In the algorithm, when deciding the next point of the search, the three acquisition functions are evaluated over the search space. 
Each acquisition function provides its own optimal point as a candidate. 
The next point at which the function is going to be evaluated is selected among the three candidates with probability given by the softmax function. The softmax function is evaluated on the cumulative reward from previous candidate points proposed by the acquisition functions, so that the strategy leans towards exploitation as the search progress. 
Once the point is selected, the Gaussian Process Regression is performed again using the updated set of data points, until the prescribed maximum number of function evaluations is reached. More details are reported in the Supplementary Material, S.2.
\section{Ensemble size and number of starting points in the test set}
\label{A:Convergence}
\begin{figure}[H]
    \centering
    \includegraphics[width=1.\textwidth]{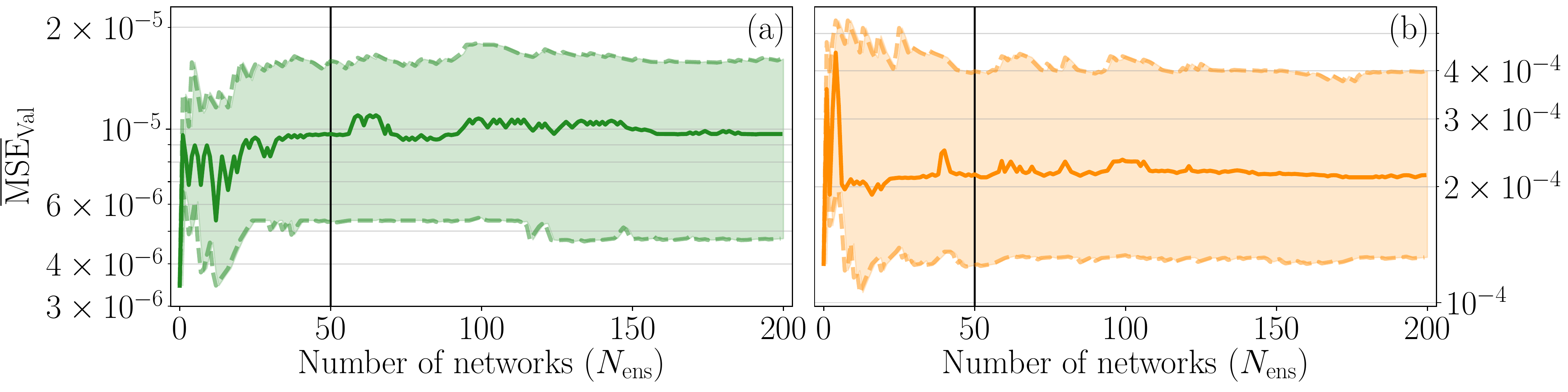}
    \caption{50th (continuous line) and 25th and 75th percentiles (dashed lines) for the Mean Squared Error in the validation set as a function of the number of networks in the ensemble in the short dataset of the Lorenz system. The hyperparameters are obtained through Bayesian Optimization in (a) chaotic Recycle Validation and (b) chaotic K-Fold Validation.
    }
    \label{ens_Conv}
\end{figure}

\begin{figure}[H]
    \centering
    \includegraphics[width=1.\textwidth]{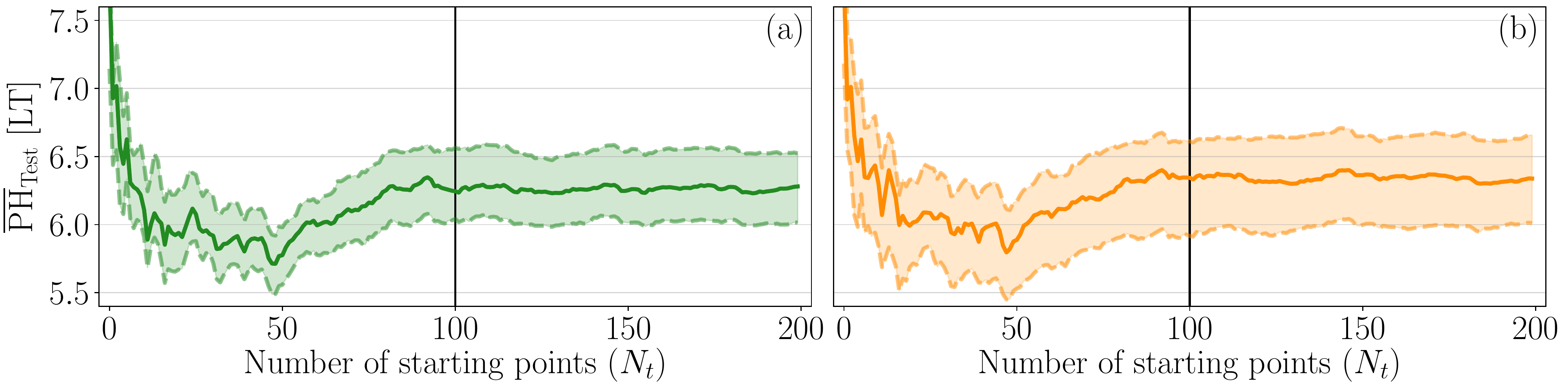}
    \caption{50th (continuous line) and 25th and 75th (dashed lines) percentiles for the Prediction Horizon in the test set for the ensemble as a function of the number of starting points in the test set in the short dataset of the Lorenz system. The hyperparameters are obtained through Bayesian Optimization in (a) chaotic Recycle Validation and (b) chaotic K-Fold Validation.
    }
    \label{Nt_Conv}
\end{figure}

First, we select the number of networks in the ensemble through the convergence of the low-order moments of the statistics of the ensemble in the validation set. 
In Fig. \ref{ens_Conv}, we show the convergence of the Mean Squared Error (MSE) in the validation set for the chaotic Recycle Validation and chaotic K-fold Validation for the Lorenz system. 
For $N_{\mathrm{ens}}=50$ networks, indicated by the vertical line, the 25th, 50th and 75th percentiles have approximately converged to their asymptotic values. 
Second, we select the number of starting points in the test set, $N_{t}$, through the convergence of the statistical properties of the ensemble in the test set. The starting points are equally spaced by 3 LTs, and start from 24 LTs in the time series of Fig. \ref{Lorenz Time Series}.
In Fig. \ref{Nt_Conv}, we show the convergence of the Prediction Horizon. 
 For $N_{t}=100$ starting points, indicated by the vertical line, the 25th, 50th and 75th percentiles have approximately converged to their asymptotic values. 
 We repeat the procedure to decide the number of starting points for the chaotic, $N_{t}=75$, and quasiperiodic, $N_{t}=50$, datasets in the Kutznetsov oscillator (results not shown). The starting points are equally spaced by 2 LTs, and start from 7.5 LTs in the time series of Fig. \ref{QP_Ch_TS}.  %

\section{Hyperparameter variations for different realizations}
\label{sec: A_fix}
\begin{figure}[H]
    \centering
    \includegraphics[width=1.\textwidth]{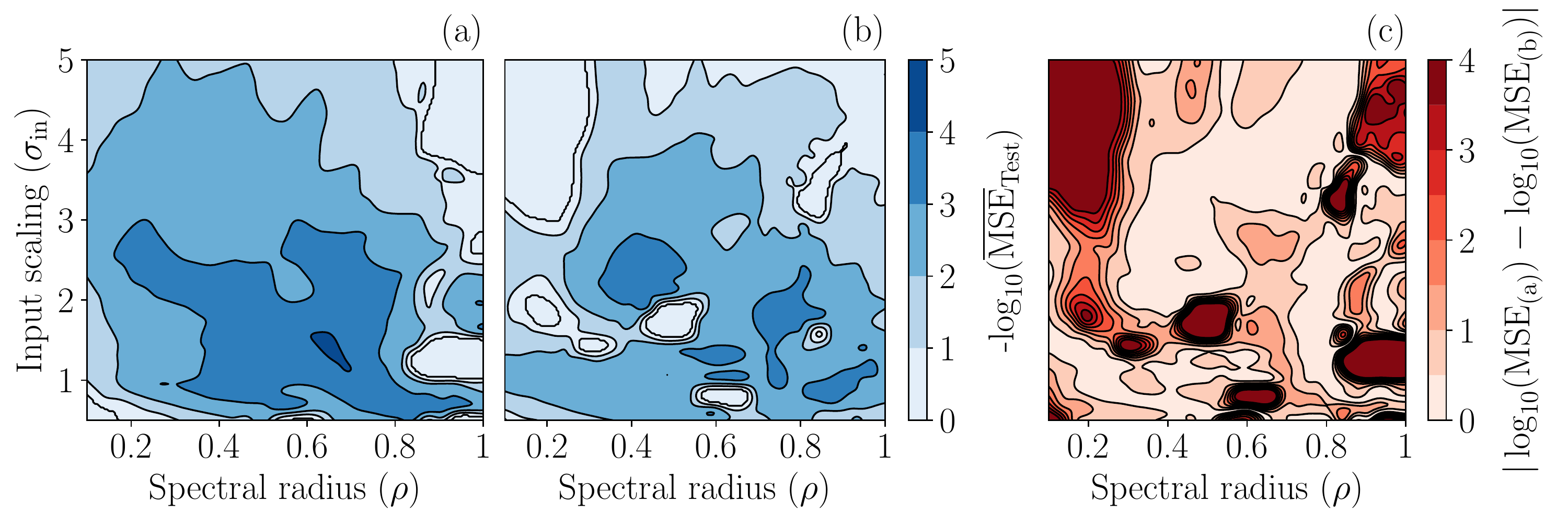}
    \caption{Mean of the Gaussian Process reconstruction of the MSE in the test set for (a,b) two representative networks in the short dataset, and (c) difference between the two networks. For visualization purposes we saturate the $\mathrm{MSE}$ to be $\leq1$ and the error to be $\leq10^4$. The Gaussian Process is based on a grid of 30$\times$30 data points.  For the same hyperparameters, the MSE can differ by orders of magnitude between the two networks.
    }
\label{Test_GP}
\end{figure}

As shown in Fig. \ref{Lack_rob}, different network realizations have different optimal hyperparameters, which vary significantly from one network realization to another other. 
This suggests that different networks need to be trained independently. 
If we select a fixed set of hyperparameters, some networks will perform poorly \cite{haluszczynski2019good}.
In this section, we quantify the difference in performance between optimizing the network independently and using a fixed set of hyperparameters for the entire ensemble. 
Figure~\ref{Test_GP} shows the mean of the Gaussian Process reconstruction of the $\log_{10}(\mathrm{MSE})$ in the test set. 
In panels (a,b), we show the MSE in the test set for two representative networks from the ensemble, while in panel (c), we show the error between the two networks. 
The two networks differ substantially. 
The same hyperparameters may result in MSEs that differ by more than four orders of magnitude.

\begin{figure}[H]
    \centering
    \includegraphics[width=1.\textwidth]{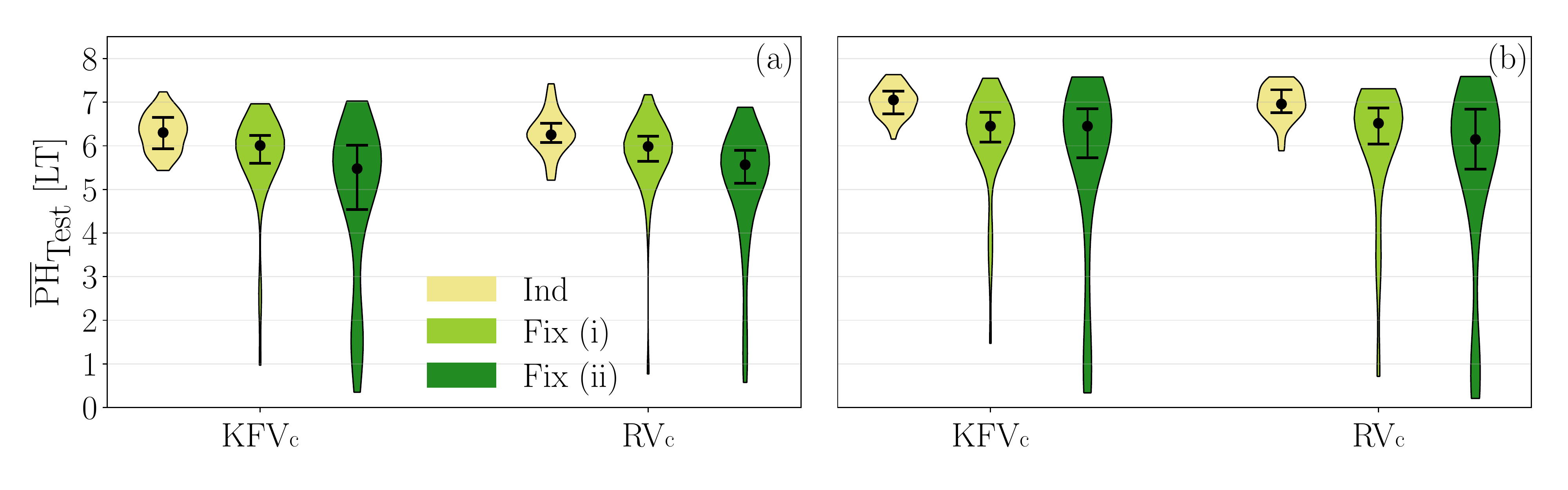}
    \caption{Violin plots and 25th (lower bar), 50th (marker) and 75th (upper bar) percentiles of the Prediction Horizon in the test set for the 50 networks ensemble in the (a) short (b) and long datasets in the Lorenz system. Independent optimization (Ind) of each network, optimal set of fixed hyperparameters (Fix (i)), and optimal hyperparameters of a single network (Fix (ii)). We use Bayesian Optimization in the chaotic K-Fold Validation (KFV$_{\mathrm{c}}$) and chaotic Recycle Validation (RV$_{\mathrm{c}}$).
    }
    \label{Lorenz_fixed_errorbars}
\end{figure}

\begin{figure}[H]
    \centering
    \includegraphics[width=1.\textwidth]{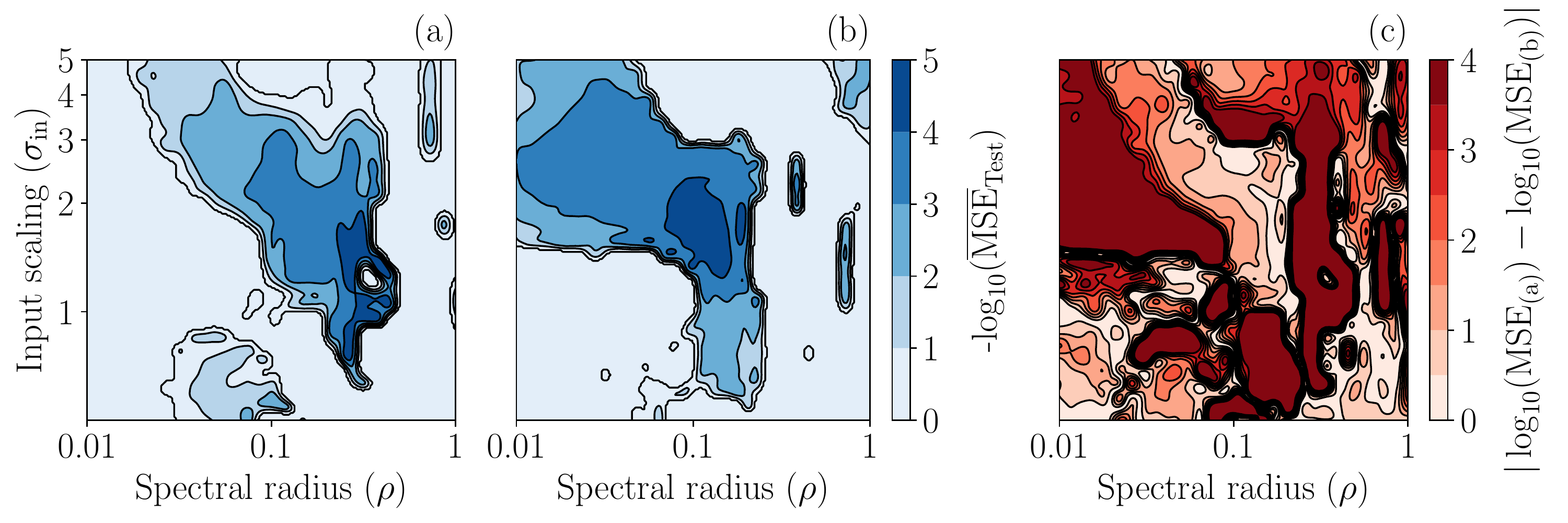}
    \caption{Mean of the Gaussian Process reconstruction of the MSE in the test set for (a,b) two representative networks in the quasiperiodic dataset, and (c) difference between the two networks. For visualization purposes we saturate the $\mathrm{MSE}$ to be $\leq1$ and the error to be $\leq10^4$. The Gaussian Process is based on a grid of 30$\times$30 data points. 
    }
\label{QP_GP}
\end{figure}

\begin{figure}[H]
    \centering
    \includegraphics[width=1.\textwidth]{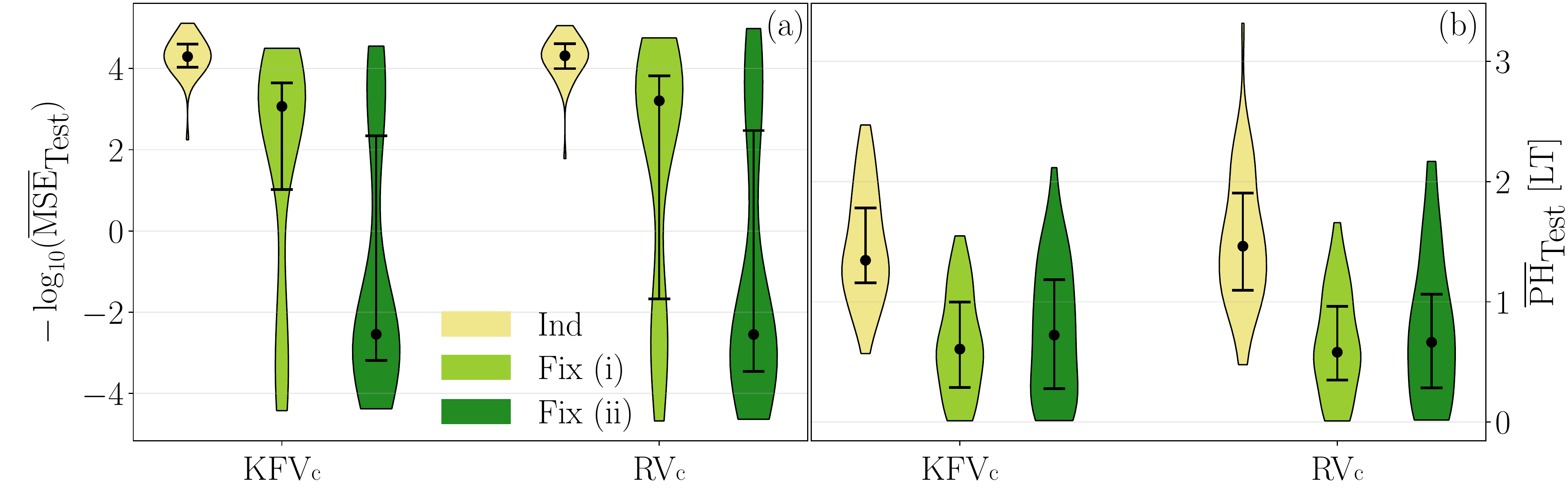}
    \caption{Violin plots and 25th (lower bar), 50th (marker) and 75th (upper bar) percentiles for the 50 networks ensemble of the MSE in the quasiperiodic dataset, (a), and the Prediction Horizon for the chaotic dataset, (b), in the test set in the Kuznetsov Oscillator. Independent optimization (Ind) of each network, optimal set of fixed hyperparameters (Fix (i)), and optimal hyperparameters of a single network (Fix (ii)). We use Bayesian Optimization in the chaotic K-Fold Validation (KFV$_{\mathrm{c}}$) and chaotic Recycle Validation (RV$_{\mathrm{c}}$).
    }
    \label{Oscillator_fixed_errorbars}
\end{figure}

To quantitatively evaluate the performance of the networks, we assess two possible choices of fixed hyperparameters: 
(i) we search the optimal fixed hyperparameters by minimizing the geometric mean over the 50 networks of the MSE in the validation set; 
(ii) we use the hyperparamters obtained by performing the search on a representative network from the ensemble and use that hyperparameters for all the networks. 
In both (i) and (ii), we perform the search using Bayesian Optimization in the chaotic K-Fold Cross Validation (KFV$_{\textrm{c}}$) and chaotic Recycle Validation (RV$_{\textrm{c}}$). 
Figure~\ref{Lorenz_fixed_errorbars} shows the violin plots and 25th, 50th and 75th percentiles for the Prediction Horizon in the test set for the Lorenz system. 
Using fixed hyperparameters yields a decrease in performance in the percentiles of around 0.5 LTs when using (i), and of more than 1 LTs when using (ii). 
In addition, the tail of the distribution prolongates to values of the Prediction Horizon below 1 LT, which means that the fixed hyperparameters perform  poorly in a fraction of the networks.
Finally, we note that the decrease in the Prediction Horizon percentiles for (ii) is larger than the improvement that we obtain when using the new validation strategies, the increased size of the dataset or the model-informed architecture. 
This means that optimizing the network independently, and therefore not using hyperparameters obtained from validating a network in another network, is key in Echo State Networks.   
Similar conclusions can be drawn for the quasiperiodic and chaotic datasets in the Kuznetsov oscillator (Fig. \ref{QP_GP},\ref{Oscillator_fixed_errorbars}). 

\newpage
\setcounter{page}{1}
\setcounter{section}{18}
\renewcommand{\thesection}{\Alph{section}}%

\end{document}